\documentclass[twocolumn,10pt]{article}

\usepackage[numbers]{natbib}
\usepackage[utf8]{inputenc}
\usepackage{pgfplots}

\usepackage{hyperref}
\usepackage{url}
\usepackage{booktabs}
\usepackage{amsfonts}
\usepackage{nicefrac}
\usepackage{microtype}

\usepackage{enumerate}
\usepackage{amsmath}
\usepackage{amsfonts}
\usepackage{amsthm}
\usepackage{amssymb}
\usepackage{bm}
\usepackage{cleveref}
\usepackage{environ}
\usepackage{subcaption}
\usepackage{stmaryrd}
\SetSymbolFont{stmry}{bold}{U}{stmry}{m}{n}

\usepackage{mathpazo}

\usepackage{wrapfig}
\usepackage{graphicx}

\usepackage{tikz}
\usetikzlibrary{plotmarks}
\usepgfplotslibrary{groupplots}

\newcommand{\x}{\ensuremath{\mathbf{x}}}
\newcommand{\z}{\ensuremath{\mathbf{z}}}
\newcommand{\w}{\ensuremath{\mathbf{d}}}
\newcommand{\G}{\ensuremath{\mathcal{G}}}
\newcommand{\R}{\ensuremath{\mathbb{R}}}
\newcommand{\bmu}{\ensuremath{\boldsymbol{\mu}}}
\newcommand{\omu}{\ensuremath{\overline{\mu}}}
\newcommand{\btheta}{{\ensuremath{\boldsymbol{\theta}}}}
\renewcommand{\u}{\ensuremath{\mathbf{u}}}
\newcommand{\eps}{\ensuremath{\varepsilon}}
\newcommand{\E}{{\ensuremath{\mathbb{E}}}}
\newcommand{\Var}{{\ensuremath{\mathbb{V}}}}
\newcommand{\CoVar}{{\ensuremath{\mathrm{Cov}}}}
\newcommand{\seq}{\ensuremath{\subseteq}}

\newtheorem{thm}{Theorem}
\newtheorem{lem}{Lemma}

\DeclareMathOperator*{\argmin}{arg\,min}

\title{Learning Implicit Generative Models Using \\ Differentiable Graph Tests}

\author{Josip Djolonga \\ ETH Z\"urich \\ \texttt{josipd@inf.ethz.ch} \and Andreas Krause \\ ETH Z\"urich \\ \texttt{krausea@ethz.ch}}

\begin{document}

\date{}
\maketitle

\begin{abstract}
	Recently, there has been a growing interest in the problem of learning rich implicit models --- those from which we can sample, but can not evaluate their density.
	These models apply some parametric function, such as a deep network, to a base measure, and are learned end-to-end using stochastic optimization.
	One strategy of devising a loss function is through the statistics of two sample tests --- if we can fool a statistical test, the learned distribution should be a good model of the true data.
	However, not all tests can easily fit into this framework, as they might not be differentiable with respect to the data points, and hence with respect to the parameters of the implicit model.
	Motivated by this problem, in this paper we show how two such classical tests, the Friedman-Rafsky and $k$-nearest neighbour tests, can be effectively smoothed using ideas from undirected graphical models -- the matrix tree theorem and cardinality potentials.
	Moreover, as we show experimentally, smoothing can significantly increase the power of the test, which might of of independent interest.
	Finally, we apply our method to learn implicit models.
\end{abstract}

\section{Introduction}
The main motivation for our work is that of learning \emph{implicit models}, i.e., those from which we can easily sample, but can not evaluate their density.
Formally, we can generate a sample from an implicit distribution $Q$ by first drawing $\z$ from some known and fixed distribution $Q_0$, typically Gaussian or uniform, and then passing it through some \emph{differentiable} function $f_\btheta$ parametrized by some vector $\btheta$ to generate $\x=f_\btheta(\z) \sim Q$.
The goal is then to optimize the parameters $\btheta$ of the mapping $\btheta$ so that $Q$ is as close as possible to some target distribution $P$,  which we can access only via iid samples.
The approach that we undertake in this paper is that of defeating statistical two-sample tests.
These tests operate in the following setting --- given two sets of iid samples, $X_1=\{\x_1, \x_2, \ldots, \x_{n_1}\}$ from $P$, and $X_2=\{\x_{n_1+1}, \x_{n_1+2}, \ldots, \x_{n_1+n_2}\}$ from $Q$, we have to distinguish between the following hypotheses
\[
	H_0 \colon P = Q \quad \textrm{ vs } \quad H_1 \colon P\neq Q.
\]
The tests that we consider start by defining a function $T \colon (\R^d)^{n_1} \times (\R^d)^{n_2} \to \R$ that should result in a \emph{low} value if the two samples come from different distributions.
Then, the hypothesis $H_0$ is rejected at significance level $\alpha\in[0,1]$ if $T(X_1, X_2)$ is lower than some threshold $t_\alpha$, which is computed using a permutation test, as explained in \Cref{sec:background}.
Going back the original problem, one intuitive approach would be to maximize
the expected statistic $\mathbb{E}_{\x_i\sim P, \z_i\sim Q_0}[T(\{\x_i\}_{i=1}^{n_1}, \{ f_\btheta(\z_i) \}_{i=n_1}^{n_1+n_2} )]$ using stochastic optimization over the parameters of the mapping $f_\btheta$.
However, this requires the availability of the derivatives $\partial T/\partial\x_i$, which is unfortunately not always possible.
For example, the Friedman-Rafsky (FR) and $k$-nearest neighbours ($k$-NN) tests, which have very desirable statistical properties (including consistency and convergence of their statistics to $f$-divergences), can not be cast in the above framework as they use the \emph{output} of a combinatorial optimization problem.
Our main contribution is the development of differentiable versions of these tests that remedy the above problem by smoothing their statistics.
We moreover show, similarly to these classical tests, that our tests are asymptotically normal under certain conditions, and derive the corresponding $t$-statistic, which can be evaluated with minimal additional complexity.
Our smoothed tests can have more power over their classical variants, as we showcase with numerical experiments.
Finally, we experimentally learn implicit models in \Cref{sec:experiments}.

\paragraph{Related work.}
The problem of two-sample testing for distributional equality has received significant interest in statistics.
For example, the celebrated Kolmogorov-Smirnov test compares two one dimensional distributions by taking the maximal difference of the empirical CDFs.
Another one-dimensional test is the runs test of \citet{wald1940test}, which has been extended to the multivariate case by \citet{friedman1979multivariate} (FR).
It is exactly this test, together with $k$-NN test originally suggested in \cite{friedman1983graph} that we analyze.
These tests have been analyzed in more detail by \citet{henze1999multivariate}, and \citet{henze1988multivariate,schilling1986multivariate} respectively.
Their asymptotic efficiency has been discussed by \citet{bhattacharya2015power}.
\citet{chen2013graph} considered the problem of tie breaking when applying the FR tests to discrete data and suggested averaging over all minimal spanning trees, which can be seen as as special case of our test in the low-temperature setting.
A very prominent test that has been more recently developed is the kernel maximum mean discrepancy (MMD) test of \citet{gretton2012kernel}, which we compare with in \Cref{sec:experiments}.
The test statistic is differentiable and has been used for learning implicit models by \citet{li2015generative,dziugaite2015training}.
\citet{sutherland2016generative} consider the problem of learning the kernel by creating a $t$-statistic using a variance estimator.
Moreover, they also pioneered the idea of using tests for model criticism --- for two \emph{fixed} distributions, one optimizes over the parameters of the test (the kernel used).
The energy test of \citet{szekely2013energy}, a special case of the MMD test, has been used by \citet{bellemare2017cramer}.

Other approaches for learning implicit models that do not depend on two sample tests have been developed as well.
For example, one approach is by estimating the log-ratio of the distributions~\cite{sugiyama2012density}.
Another approach, that has recently sparked significant interest, and can be also seen as estimating the log-ratio of the distributions, are the generative adversarial networks (GAN) of \citet{goodfellow2014generative}, who pose the problem as a two player game.
One can, as done in \cite{sutherland2016generative}, combine GANs with two sample tests by using them as feature matchers at some layer of the generating network~\cite{salimans2016improved}.
\citet{nowozin2016f} minimize an arbitrary $f$-divergence~\cite{ali1966general} using a GAN framework, which can be related to our approach, because the limit of our tests converge to specific $f$-divergences, as explained in \Cref{sec:background}.
For an overview of various approaches to learning implicit models we direct the reader to \citet{mohamed2016learning}.
\section{Classical Graph Tests}\label{sec:background}
Let us start by introducing some notation.
For any set $X=\{\x_1, \x_2, \ldots, \x_n\}$ of points in $\R^d$, we will denote by $\G(X)=(X,E)$ the \emph{complete directed graph}\footnote{For the FR test we will arbitrarily choose one of the two edges for each pair of nodes.} defined over the vertex set $X$ with edges $E$.
We will moreover weigh this graph using some function $d \colon \R^d\times \R^d \to [0,\infty)$, e.g.\ a natural choice would be $d(\x,\x')=\|\x-\x'\|$.
Similarly, we will use $d(e)$ for the weight of the edge $e$ under $d(\cdot, \cdot)$.
For any labelling of the vertices $\pi : X\to \{1,2\}$, and any edge $e\in E$ with adjacent vertices $i$ and $j$ we define\footnote{We use the Iverson bracket $\llbracket S \rrbracket$ that evaluates to 1 if $S$ is true and 0 otherwise.} $\Delta_\pi(e)=\llbracket \pi(i)\neq \pi(j)\rrbracket$, i.e., $\Delta_\pi(e)$ indicates if its end points of $e$ have \emph{different} labels under $\pi$.
Remember that we are given $n_1$ points $X_1=\{\x_1, \x_2, \ldots, \x_{n_1}\}$ from $P$, and $n_2$ points $X_2=\{\x_{n_1+1}, \x_{n_1+2}, \ldots, \x_{n_1+n_2}\}$ from $Q$.
In the remaining of the paper we will use $n=n_1+n_2$ for the total number of points.
The tests are based on the following four-step strategy.
\begin{enumerate}[(i)]
	\item Pool the samples $X_1$ and $X_2$ together into $X=X_1\cup X_2 = \{\x_1, \x_2, \ldots, \x_{n_1+n_2}\}$, and create the graph $\G(X)$.
	      Define the mapping $\pi^* \colon X \to \{1,2\}$ evaluating to 1 on $X_1$ and to 2 on $X_2$.
	\item Using some well-defined algorithm $\mathcal{A}$ choose a subset $U^*=\mathcal{A}(\G(X))$ of the \emph{edges} of this graph with the underlying motivation that it defines some neighbourhood structure.
	\item Count how many edges in $U^*$ connect points from $X_1$ with points from $X_2$, i.e., compute the statistic $T_{\pi^*}(U^*) = \sum_{e\in U^*}  \Delta_{\pi^*}(e)$.
	\item Reject $H_0$ for small values of $T_{\pi^*}(U^*)$.
\end{enumerate}
These tests condition on the data and are executed as permutation tests, so that the critical value in step (iv) is computed using the quantiles of $\E_{\pi\sim H_0}T_\pi(U^*)$, where $\pi \colon X\to\{1,2\}$ is drawn uniformly at random from the set of ${n_1 + n_2\choose n_1}$ labellings that map exactly $n_1$ points from $X$ to 1.
Formally, the $p$-value is given as $\mathbb{E}_{\pi\sim H_0}[\llbracket T_{\pi^*}(U^*)\geq T_\pi(U^*)\rrbracket]$.
We are now ready to introduce the two tests that we consider in this paper, which are obtained by using a different neighbourhood selection algorithm $\mathcal{A}$ in step (ii).

\paragraph{Friedman-Rafsky (FR).}
This test, developed by \citet{friedman1979multivariate}, uses the minimum-spanning tree (MST) of $\G(X)$ as the neighbourhood structure $U^*$, which can be computed using the classical algorithms of \citet{prim1957shortest} and \citet{kruskal1956shortest} in time $O(n^2\log n)$.
If we use $d(\x_i,\x_j)=\|\x_i-\x_j\|$, the problem is also known as the Euclidean spanning tree problem, and in this case \citet{henze1999multivariate} have proven that the test is consistent and has the following asymptotic limit.
\begin{thm}[\cite{henze1999multivariate}]
	If $d(\x,\x')=\|\x-\x'\|$ and $n_1 / (n_1 + n_2)\to \alpha\in(0,1)$, then it almost surely holds that
	\[
		\frac{T_{\pi^*}(U^*)}{n_1+n_2} \to 2\alpha(1-\alpha) \int \frac{p(\x)q(\x)}{\alpha p(\x) + (1-\alpha) q(\x)}d\x,
	\]
	where $p$ and $q$ are the densities of $P$ and $Q$.
\end{thm}
As noted by \citet{berisha2015empirical}, after some algebraic manipulation of the right hand side of the above equation, we obtain that $1-T_{\pi^*}(U^*)\frac{n_1+n_2}{2n_1n_2}$ converges almost surely to the following $f$-divergence~\cite{ali1966general}
\begin{align*}
	D^{\textrm{FR}}_\alpha(P \,\|\, Q) & = \frac{1}{4\alpha(1-\alpha)}\int \frac{(\alpha p(\x) - (1-\alpha) q(\x))^2}{\alpha p(\x) + (1-\alpha)q(\x)} d\x \\
	                                   & - \frac{(2\alpha-1)^2}{4\alpha(1-\alpha)}.
\end{align*}
In \cite{berisha2015empirical} it is also noted that if $n_1=n_2$, then $\alpha=1/2$ and in that case $D_{1/2}$ is equal to $2\int\frac{(p(\x)-q(\x))^2}{p(\x)+q(\x)}d\x$, which is known as the symmetric $\chi^2$ divergence.

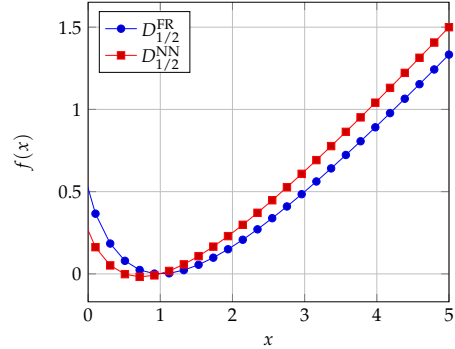
\begin{figure}[t]
	\centering
	\scalebox{0.7}{
		\begin{tikzpicture}
			\begin{axis}[
					xlabel={$x$},
					samples=50,
					ylabel={$f(x)$},
					xmin=0, xmax=5,
					grid,
					legend pos=north west,
				]
				\addplot {(0.5 * \x - 0.5)^2 / (0.5 * \x + 0.5)};
				\addlegendentry{$D^\textrm{FR}_{1/2}$};
				\addplot {(0.25*\x^2 + 0.5) / (0.5 * \x + 0.5) - 0.75};
				\addlegendentry{$D^\textrm{NN}_{1/2}$};
			\end{axis}
		\end{tikzpicture}
	}

	\caption{\footnotesize The functions generating the $f$-divergences.}
	\label{fig:fdiv}
\end{figure}
\paragraph{$k$-nearest-neighbours ($k$-NN).}
Maybe the most intuitive way to construct a neighbourhood structure is to connect each point $\x_j\in X$ to its $k$ nearest neighbours.
Specifically, we will add the edge $\x_i\to\x_j$ to $U^*$ iff $\x_i$ is one of the $k$ closest neighbours of $\x_j$ as measured by $d(\x,\x')$.
If one uses the Euclidean norm, then the asymptotic distribution and the consistency of the test have been proven by \citet{schilling1986multivariate}.
These results has been extended to arbitrary norms by \citet{henze1988multivariate}, who also proved the limiting behaviour of the statistic as $n\to\infty$.
\begin{thm}[\cite{henze1988multivariate}]
	If $n_1 / (n_1 + n_2)\to \alpha\in(0,1)$, then $1 - \frac{T_{\pi^*}(U^*)}{(n_1+n_2)k}$ converges in probability to
	\[
		D^{\textrm{NN}}_\alpha (P \,\|\, Q)\equiv \int \frac{\alpha^2 p^2(\x) + (1-\alpha)^2q^2(\x)}{\alpha p(\x) + (1-\alpha) q(\x)}d\x,
	\]
	where $p$ and $q$ are the continuous densities of $P$ and $Q$.
\end{thm}
As for the FR test, we can also re-write the limit as an $f$-divergence\footnote{This $f$ does not vanish at one, but we can simply shift it.} corresponding to $f(t)=(\alpha^2 t^2 + (1-\alpha)) / (\alpha t + (1-\alpha))$.
Moreover, if we compare the integrands in $D^{\textrm{FR}}_\alpha$ and $D^{\textrm{NN}}_\alpha$, we see that they are related and they differ by the term $2\alpha(1-\alpha)p(\x)q(\x)$ in the numerator.
The fact that they are closely related can be also seen from \Cref{fig:fdiv}, where we plot the corresponding $f$-functions for the $n_1=n_2$ case.
\section{Differentiable Graph Tests}
While the tests from the previous section have been studied from a statistical perspective, we can not use them to train implicit models because the derivatives $\partial T / \partial \x_i$ are either zero or do not exist, as $T$ takes on finitely many values.
The strategy that we undertake in this paper is to \emph{smooth} them into continuously differentiable functions by relaxing them to expectations in natural probabilistic models.
To motivate the models we will introduce, note that for both the $k$-NN and the FR test, the optimal neighbourhood is the solution to the following optimization problem
\begin{equation}\label{prob:uopt}
	U^* = \argmin_{U\seq E} \sum_{e\in U} d(e) \; \textrm{s.t.\ } \nu(U)=1,
\end{equation}
where  $\nu \colon 2^E \to \{0,1\}$ indicates if the set of edges is \emph{valid}, i.e., if every vertex has exactly $k$ neighbours in the $k$-NN case, or if the set of edges forms a poly-tree in the MST case.
Moreover, note that once we fix $n_1$ and $n_2$, the optimization problem \eqref{prob:uopt} depends only on the edge weights $d(e)$, which we will concatenate in an arbitrary order and store in the vector $\w\in\R^{|E|}$.
We want to design a probability distribution over $U$ that focuses on those configurations $U$ that are both feasible and have a low cost for problem~\eqref{prob:uopt}.
One such natural choice is the following Gibbs measure
\begin{equation}\label{eqn:model}
	P(U \mid \w/\lambda) = e^{- \sum_{e\in U} d(e)/\lambda-A(-\w/\lambda)}\nu(U),
\end{equation}
where $\lambda$ is the so-called temperature parameter, and $A(-\w/\lambda)$ is the log-partition function that ensures that the distribution is normalized.
Note that $U^*$ is a MAP configuration of this distribution~\eqref{eqn:model}, and the distribution will concentrate on the MAP configurations as $\lambda\to 0$.
Once we have fixed the model, the strategy is clear --- replace the statistic $T_{\pi^*}(U^*)$ with its expectation $\E_U[T_{\pi^*}(U)]$, which results in the following smooth statistic
\begin{align*}\label{eqn:smooth}
	T_{\pi*}(U^*) \longrightarrow  T_{\pi^*}^{\lambda} & \equiv \mathbb{E}_{U\sim P(\cdot \mid \w, \lambda)}[T_{\pi^*}(U)] \\
	                                                   & = \sum_{e\in E} \Delta_{\pi^*}(e) \bmu(\w/\lambda)_e,
\end{align*}
where $\bmu(\w/\lambda)$ are the marginal probabilities of the edges, i.e., $[\bmu(\w/\lambda)]_e = \mathbb{E}_{P(U \mid \w / \lambda)}[\llbracket e\in U\rrbracket]$.
Hence, we can compute the statistic as long as we can perform inference in \eqref{eqn:model}.
To compute its derivatives we can use the fact that \eqref{eqn:model} is a member of the exponential family.
Namely, leveraging the classical properties of the log-partition function \cite[Prop.~3.1]{wainwright2008graphical}, we obtain the following identities
\begin{equation}\label{eqn:grads}
	\begin{aligned}
		\bmu(\w/\lambda) & = \nabla A(-\w/\lambda), \textrm{ and}                                       \\
		\frac{\partial \bmu(\w/\lambda)_e}{\partial \bmu(\w/\lambda)_{e'}}
		                 & = \mathbb{E}_{P(U \mid \w / \lambda)}[\llbracket \{e, e'\} \seq U\rrbracket]
		\\
		                 & - \bmu(\w/\lambda)_e\bmu(\w/\lambda)_{e'}.
	\end{aligned}
\end{equation}
Thus, if we can compute both first- and second-order moments under \eqref{eqn:model}, we get both the smoothed statistic and its derivative.
We show how to do this for the $k$-NN and FR tests in \Cref{sec:computation}.

\paragraph{A smooth $p$-value.}
Even though one can directly use the smoothed test statistic $T_{\pi^*}^\lambda$ as an objective when learning implicit models, it does not necessarily mean that lower values of this statistic result in higher $p$-values.
Remember that to compute a $p$-value, one has to run a permutation test by computing quantiles of $T_{\pi}^\lambda$ under random draws of the permutation $\pi\sim H_0$.
However, as this procedure is not smooth and can be costly to compute, we suggest as an alternative that does not suffer from these problems the following $t$-statistic
\begin{equation}\label{eqn:tstat}
	t^{\lambda}_{\pi^*} = \frac{T_{\pi^*}^{\lambda} - \mathbb{E}_{\pi\sim H_0}[T_{\pi}^{\lambda}]}{\sqrt{\mathbb{V}_{\pi\sim H_0}[T_{\pi}^{\lambda}]}}.
\end{equation}
The same strategy has been undertaken for the FR and $k$-NN tests in \cite{friedman1979multivariate,henze1999multivariate,schilling1986multivariate}.
Before we show to compute the first two moments under $H_0$, we need to define the matrix $\Pi$ holding the second moments of the variables $\Delta_{\pi}(e)$.
\begin{lem}[\cite{friedman1979multivariate}]
	The matrix $\Pi\in\R^{|E|\times |E|}$ with entries $\Pi_{e,e'}=\E_{\pi\sim H_0}[\Delta_\pi(e)\Delta_\pi(e')]$ is equal to
	\[
		\Pi_{e, e'} =
		\begin{cases}
			\frac{2 n_1n_2}{n(n-1)}                        & \textrm{ if }  \delta(e)=\delta(e'), \textrm{ or}         \\
			\frac{n_1n_2}{n(n-1)}                          & \textrm{ if }  |\delta(e)\cap \delta(e')|=1, \textrm{ or} \\
			\frac{4n_1n_2(n_1-1)(n_2-1)}{n(n-1)(n-2)(n-3)} & \textrm{ if }  \delta(e) \cap \delta(e') = \emptyset,
		\end{cases}
	\]
	where $\delta(e)$ is the set of vertices incident to the edge $e\in E$.
\end{lem}
\begin{thm}\label{thm:moments}
	Assume that all valid configurations $U$ satisfy $|U|=m$, i.e.\ that $\nu(U)\neq 0$  implies $|U|=m.$\footnote{Note that we have $m=kn$ for $k$-NN and $m=n-1$ for FR.}
	Then, the first two moments of the statistic under $H_0$ are
	\begin{align*}
		\E_{\pi\sim H_0}[T^\lambda_{\pi^*}]   & = 2 m n_1n_2/n(n-1), \textrm{ and}                                              \\
		\Var_{\pi\sim H_0}[T^\lambda_{\pi^*}] & = \bmu(\w/\lambda)^T\Pi \bmu(\w/\lambda)-  4 \frac{n_1^2 n_2^2}{n^2(n-1)^2}m^2.
	\end{align*}
\end{thm}
While the computation of the mean is trivial, it seems that the computation of the variance needs $O(|E|^2)$ operations.
However, we can simplify its computation to $O(|E|)$ using the following result.
\begin{lem}\label{lem:simple-var}
	Define $\chi_1=\frac{n_1n_2}{n(n-1)}$ and $\chi_2=\frac{4(n_1-1)(n_2-1)}{(n-2)(n-3)}$.
	Then, the variance can be computed as
	\begin{align*}
		\sigma^2 & = \chi_1(1-\chi_2)\sum_v (\sum_{e\in \delta(v)}\mu_e)^2                   \\
		         & + \chi_1\chi_2\sum_{e \| e'}\mu_e\mu_{e'} + \chi_1(\chi_2 - 4\chi_1) m^2,
	\end{align*}
	where $\sum_{e \| e}$ sums over all pairs of parallel edges, i.e., those connecting the same end-points.
\end{lem}

\paragraph{Approximate normality of $t_{\pi^*}^\lambda$.}
To better motivate the use of a $t$-statistic, we can, similarly to the arguments in \cite{friedman1979multivariate,henze1999multivariate,schilling1986multivariate}, show that it is is close to a normal distribution by casting it as a \emph{generalized correlation coefficient}~\cite{daniels1944relation,friedman1983graph}.
Namely, these are tests whose statistics are the form form $\kappa = \sum_{i=1}^n \sum_{j=1}^n \omu_{i,j}b_{i,j}$, and whose critical values are computed using the distribution of $\sum_{i=1}^n \sum_{j=1}^n \omu_{i,j}b_{\pi(i),\pi(j)}$, where $\pi$ is a random permutation on $\{1, 2, \ldots, n\}$.
It is easily seen that we can fit the suggested tests in this framework if we set $\omu_{i,j}=\frac{1}{2}(\bmu(\w/\lambda)_{i\to j} + \bmu(\w/\lambda)_{j\to i})$ and $b_{i,j}=\Delta_{\pi^*}(\{i,j\})$.
Then, using the conditions of \citet{barbour1986random}, we obtain the following bound on the deviation from normality.

\begin{thm}\label{thm:normality}
	Let $n_1 / (n_1 + n_2)\to\alpha\in (0, 1)$, and define
	\begin{itemize}
		\item $S_2 = \sum_{i,j,k} \omu_{i,j} \omu_{i,k}$, i.e., the expected number of edges sharing a vertex,
		\item $S_3 = \sum_{i,j,k,m} \omu_{i,j} \omu_{i,k}\omu_{i,m}$, i.e., the expected number of 3 stars, and
		\item $L_4 = \sum_{i,j,k,m} \omu_{i,j} \omu_{j,k}\omu_{k,m}$, i.e., the expected number of paths with 4 nodes.
	\end{itemize}
	Then, the Wasserstein distance between the permutation null $\E_{\pi\sim H_0}[T_{\pi}^{\lambda}(U^*)]$ and the standard normal is of order $O\big( ( nk^3 +kS_2+S_3+L_4) / \sigma^3\big)$.
\end{thm}

Let us analyze the above bound in the setting that we will use it --- when $n_1=n_2$.
First, let us look at the variance, as formulated in \Cref{lem:simple-var}.
The last term can be ignored as it is always non-negative because $\chi_2\geq 4\chi_1$ (shown in the appendix).
Because $\sum_{e\in\delta(v)}\mu_e\geq 1$, it follows that the variance grows as $\Omega(n)$.
Thus, without any additional assumption on the growth of the neighbourhoods, we have asymptotic normality as $n\to\infty$ if the numerator is of order $o(n^{1.5})$.
For example, that would be satisfied if the largest neighbourhood $\max_i \sum_{e\in\delta(i)} \omu_e$ grows as $o(n^{1/6})$.
Note that in the low temperature setting (when $\lambda\to 0$), the coordinates of $\bmu$ will be very close to either zero or one.
As observed by~\citet{friedman1979multivariate}, in this case $S_2=O(1)$ as the nodes of both the $k$-NN and MST graphs have nodes whose degree is bounded by a constant independent of $n$ as $n\to\infty$~\cite{yukich2006probability}.
We also observe experimentally in \Cref{sec:experiments} that the distribution gets closer to normality as $\lambda$ decreases.
\section{The Differentiable $k$-NN and Friedman-Rafsky Tests}\label{sec:computation}
In this section, we discuss these two tests in more detail and show to efficiently compute their statistics.
Remember that to compute and optimize \emph{both} $T_{\pi^*}^\lambda$ and $t_{\pi^*}^\lambda$ we have to be able to perform inference in the model $P(U) = \exp(-\sum_e d(e)/\lambda - A(-\w/\lambda))\nu(U)$, by computing the first and- second-order moments of the edge indicator variables.
We would stress that, in the learning setting that we consider $n$ refers to the number of data-points in a \emph{mini-batch}.

\paragraph{$k$-NN.} The constraint $\nu(\cdot)$ in this case requires the total number of edges in $U$ incoming at each node to be exactly $k$.
First, note that the problem completely \emph{separates} per node, i.e., the marginals of edges with different target vertices are independent.
Formally, if we denote by $U_i$ the set of edges incoming at vertex $i$, then $U_i$ and $U_j$ are independent for $i\neq j$.
Hence, for each node $i$ \emph{separately}, have to perform inference in
\[
	P(U_i) \propto \exp(-\sum_{j\in U_i} d(\x_i, \x_j) / \lambda )\llbracket |U_i|=k\rrbracket,
\]
which is a special case of the cardinality potentials considered by \citet{tarlow2012fast,swersky2012cardinality}.
\citet{swersky2012cardinality} consider the same model, and note that we can compute \emph{all marginals} in time $O(nk)$ using the algorithm in \cite{tarlow2012fast}, which works by re-writing the model as a chain CRF and running the classical forward-backward algorithm.
Hence, the total time complexity to compute the vector $\bmu(\w/\lambda)$ is $O(n^2k)$. Moreover, as marginalization requires only simple operations, we can compute the derivatives with any automatic differentiation software, and we thus do not provide formulas for the second-order moments.
In \cite{swersky2012cardinality} the authors provide an approximation for the Jacobian, which we did not use in our experiments, but instead we differentiate through the messages of the forward-backward algorithm.

As a concrete example, let us work out the simplest case --- the $k$-NN test with $k=1$.
In this case, the smoothed statistic reduces to
\[
	T^{\lambda}_{\pi^*}(\x_1, \ldots, \x_n) =
	\sum_{i=1}^n \sum_{\substack{j=1 \\ \pi^*(i)\neq \pi^*(j)}}^n s_i(\x_1, \ldots, \x_n)_j,
\]
where $s_i(\x_1,\ldots,\x_n)=\texttt{softmax}(-\otimes_{l\neq i}\|\x_i-\x_l\|/\lambda)$.
In other words, for each $i$ you compute the \texttt{softmax} of the distances to all other points using $s_i$, and then sum up only those positions that correspond to points from the \emph{other} sample.
One interpretation of the loss is the following --- maximize the number of \emph{incorrect} predictions if we are to estimate the label $\pi(i)$ from $\x_i$ using a soft $1$-nearest neighbour approach.

Furthermore, we can also make a clear connection between the smooth $1$-NN test and neighbourhood component analysis (NCA)~\cite{goldberger2005neighbourhood}. Namely, we can see NCA as learning a mapping $h \colon \x\to A\x$ so that the test \emph{distinguishes} (by minimizing $T^\lambda_{\pi^*}$) the two samples as best as possible after applying $h$ on them.
The extension of NCA to $k$-NN~\cite{tarlow2013stochastic} can be also seen as minimizing the test statistic for a particular instance of their loss function.

\paragraph{Friedman-Rafsky.} The model that we have to perform inference in for this test seems extremely complicated and intractable at first because the constraint has the form
$\nu(U) = \llbracket U \textrm{ forms a spanning tree}\rrbracket$.
First, note that if $\w/\lambda$ had all entries equal to a constant $\gamma$, we have that $A(-\w/\lambda)=(1-n)\gamma + \log c_{G(X)}$, where $c_{\mathcal{G}(X)}$ is the number of spanning trees in the graph $\mathcal{G}(X)$, and can be computed using Kirchoff's (also known as the matrix-tree) theorem.
To treat the weighted case, we use the approach of \citet{lyons2003determinantal}, who has showed that the above model is a determinantal point process (DPP), so that marginalization can be done exactly as follows.
First, create the incidence matrix $A\in\{-1,0,+1\}^{(n-1)\times |E|}$ of the graph $\mathcal{G}(X)$ after removing an arbitrary vertex, and construct its Laplacian $L=A\texttt{diag}\big[\exp(-\w/\lambda)\big]A^T$.
Then, if we compute $H=L^{-1/2}A\texttt{diag}\big[\exp(-\w/(2\lambda))\big]$, the distribution $P(U)$ is a DPP with kernel matrix $K=H^TH$, implying that for every $W\subseteq E$
\[
	\mathbb{E}_{P(U\mid \w/\lambda)}[\llbracket W\subseteq U\rrbracket] =
	\det K_W,
\]
where $K_W$ is the $|W|\times|W|$ submatrix of $K$ formed by the rows and columns indexed by $W$.
Thus, we can easily compute all marginals and  the smoothed test statistic and its derivatives using \eqref{eqn:grads} as
\begin{align*}
	\mu_{i\to j}                                        & = e^{-d(\x_i,\x_j)/\lambda}(\u_i - \u_j)^TL^{-1}(\u_i-\u_j), \textrm{ and}                 \\
	\frac{\partial \mu_{i\to j}}{\partial \mu_{k\to l}} & = e^{-\frac{d(\x_i,\x_j) + d(\x_k, \x_l)}{\lambda}}((\u_i - \u_j)^TL^{-1}(\u_k - \u_l))^2,
\end{align*}
where $\u_i$ is the vector with coordinates equal to zero, except the $i$-th coordinate which is one.
Note that if we first compute the inverse $L^{-1}$, all quantities of the form $L^{-1}(\u_i - \u_j)$ can be computed in time $O(n)$ as the vectors $\u_i$ have a single non-zero entry, for a total complexity of $O(n^3)$.

To speed up this computation we can leverage the existing theory on fast solvers of Laplacian systems.
Let us first create from $\G(X)$ the graph $e^{\G}(X)$ that has the same structure as $\G(X)$, but with edge weights $e^{-d(e)/\lambda}$ instead of $d(e)$.
Hence, in this graph, a large weight between $\x$ and $\x'$ indicates that these two points are \emph{similar} to one another.
In $e^{\G}(X)$, the marginals $\bmu_e$ are also known as \emph{effective resistances}\footnote{For additional properties of the effective resistances see~\cite{chandra1996electrical}.}.
\citet{spielman2011graph} provide a method to compute \emph{all} marginals at once in time that is $\tilde{O}(rn^2/\eps^2)$, where $\eps$ is the desired relative precision and $r=\frac{1}{\lambda}(\max_e d(e) - \min_e d(e))$.
The idea is to first solve for $Z^T=L^{-1}A\texttt{diag}\big[\exp(-\w/2\lambda)\big]R$ where $R\in\{-1/\sqrt{k}, +1/\sqrt{k}\}^{|E|\times p}$ is a random projection matrix with elements chosen uniformly from $\{-1/\sqrt{k}, +1/\sqrt{k}\}$ and $p=O(\log n/\eps^2)$.
Then, the suggested approximation is $\mu_{i\to j} \approx \|Z(\u_i-\u_j)\|^2$.
While computing $Z$ na\"ively would take $O(n^3 + n^2p)$, one achieves the claimed bound with the Laplacian solver of \citet{spielman2014nearly}.

As an extra benefit, the above connection provides an alternative interpretation of the smoothed FR test.
Namely, assume that we want to create a \emph{spectral sparsifier}~\cite{spielman2011spectral} of $e^{\G}(X)$, which should contain \emph{significantly less} edges, but be a good summary of the graph by having a similar spectrum.
\citet{spielman2011graph} provide a strategy to create such a sparsifier by sampling edges randomly, where edge $e$ is sampled proportional to $\mu_e$.
Hence, by optimizing $T^{\lambda}_{\pi^*}$ we are encouraging the constructed sparsifier of $e^{\G}(X)$ to have in expectation as many edges as possible connecting points from $X_1$ with points from $X_2$.
\section{Experiments}\label{sec:experiments}

We implemented our methods in Python using the \texttt{PyTorch} library.
For the $k$-NN test, we have adapted the code accompanying~\cite{swersky2012cardinality}.
Throughout this section we used a 10 dimensional normal as $Q_0$, drew samples of equal size $n_1=n_2$, and used the $\ell_2$ norm $d(\x,\x')=\|\x-\x'\|_2$ as a weighting function.
We provide additional details in \Cref{app:experiments}.

\begin{figure*}
	\centering

	\begin{subfigure}[t]{\linewidth}
		{\footnotesize
			\includegraphics[width=0.32\linewidth]{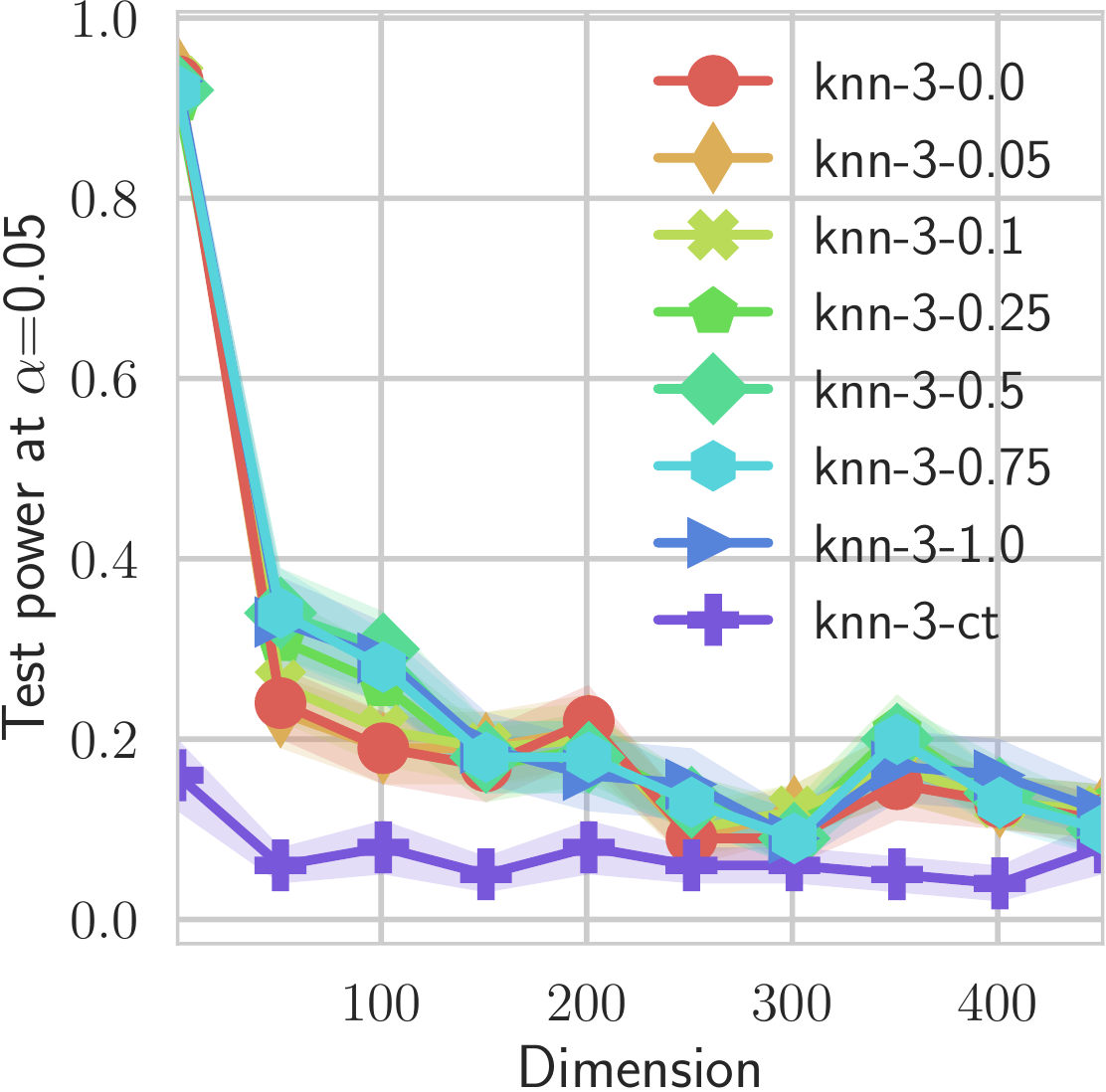}
			\includegraphics[width=0.32\linewidth]{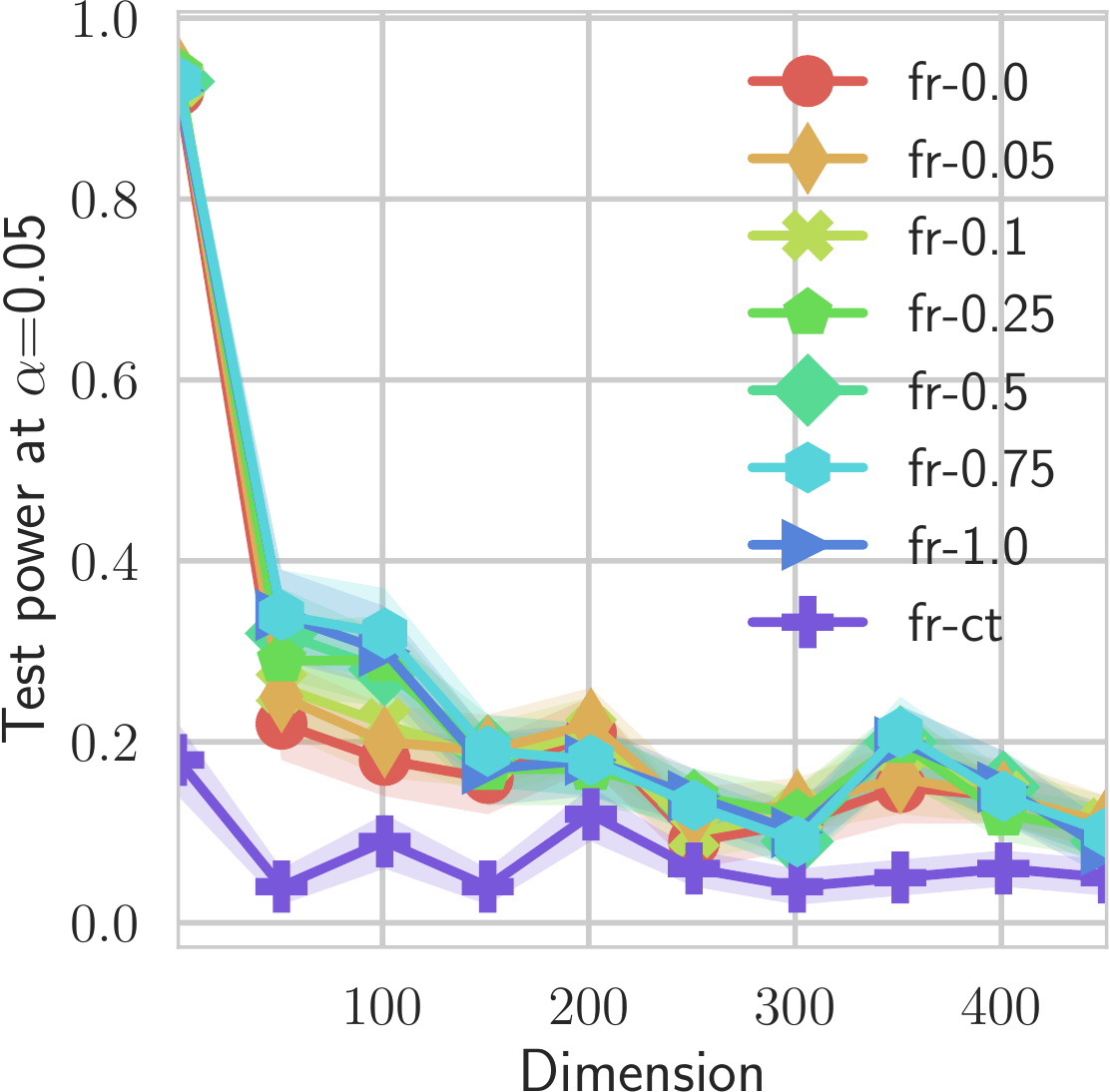}
			\includegraphics[width=0.32\linewidth]{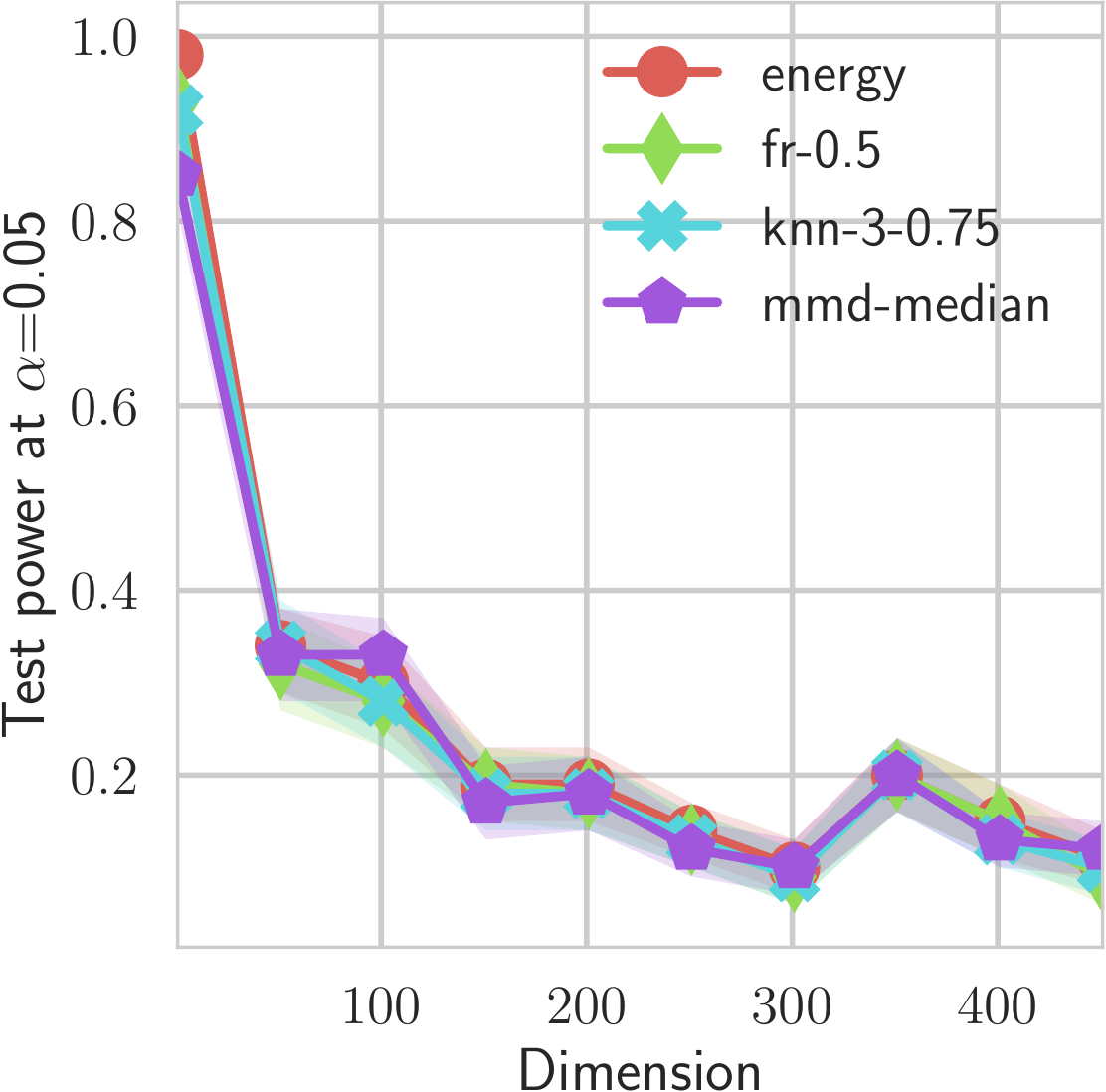}
		}
		\caption{\footnotesize Power against the alternative $(\mu=0.5, \sigma=1)$ from $n_1=n_2=128$ samples.}

	\end{subfigure}

	\begin{subfigure}[t]{\linewidth}
		{\footnotesize
			\includegraphics[width=0.32\linewidth]{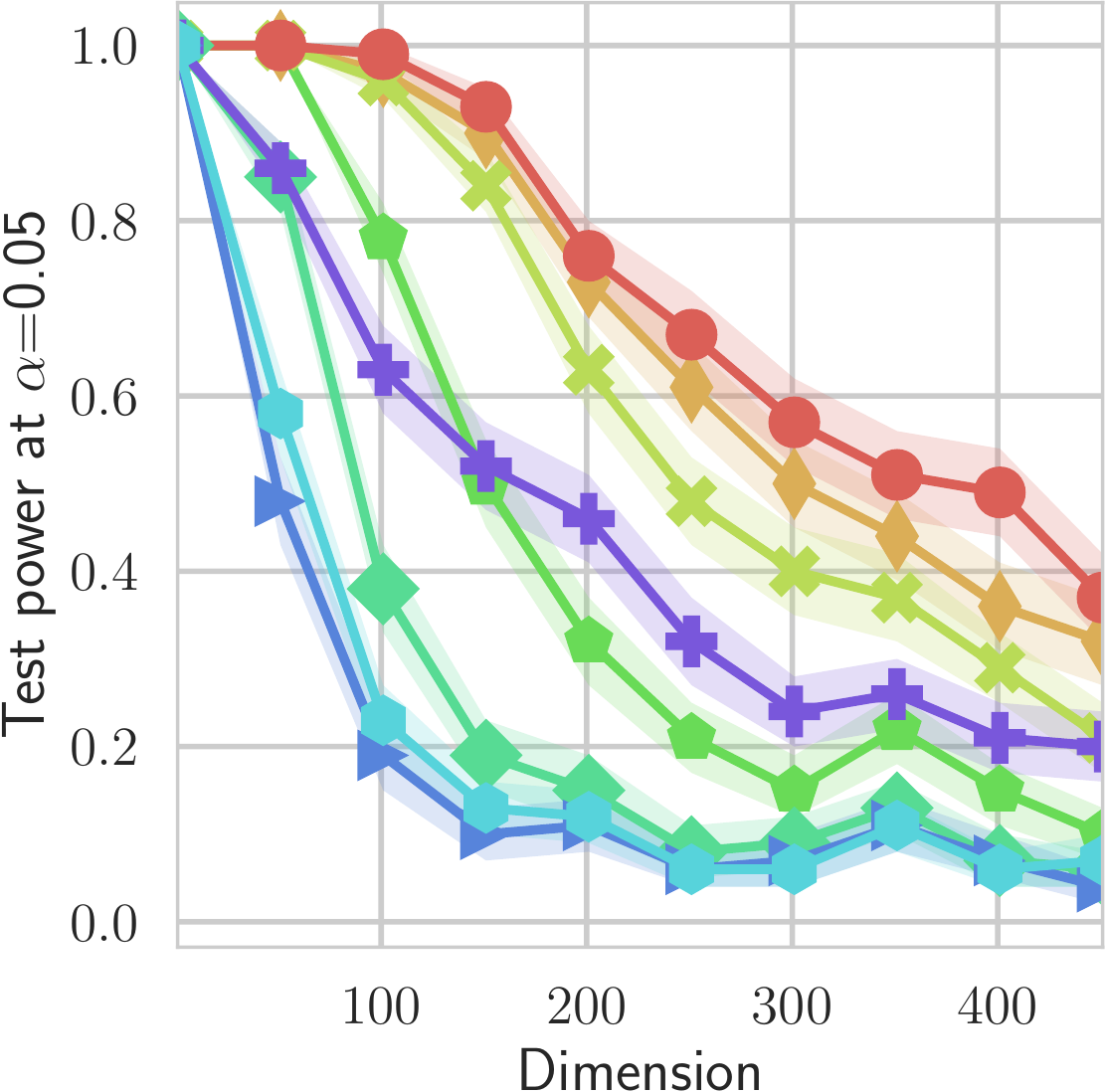}
			\includegraphics[width=0.32\linewidth]{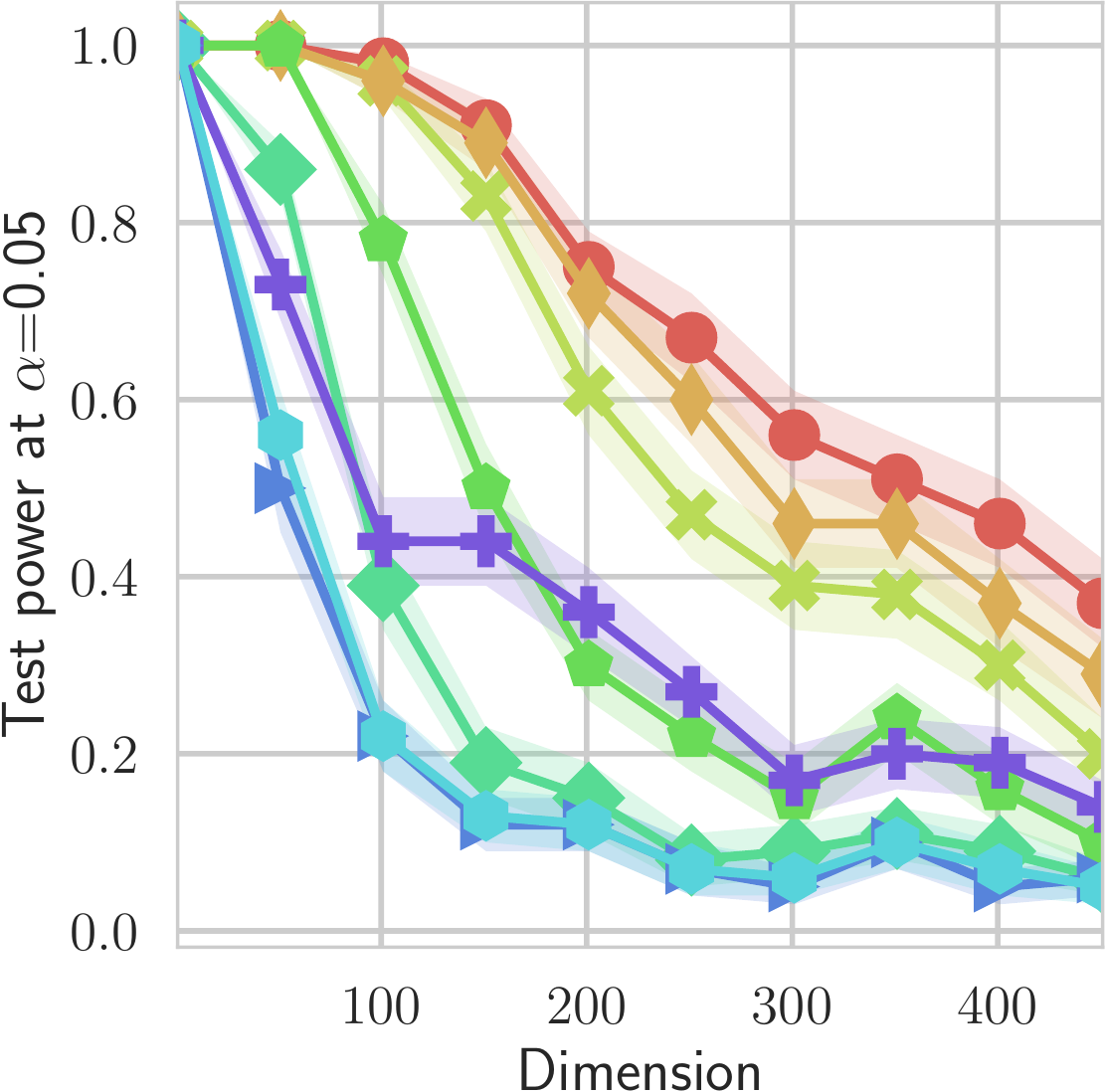}
			\includegraphics[width=0.32\linewidth]{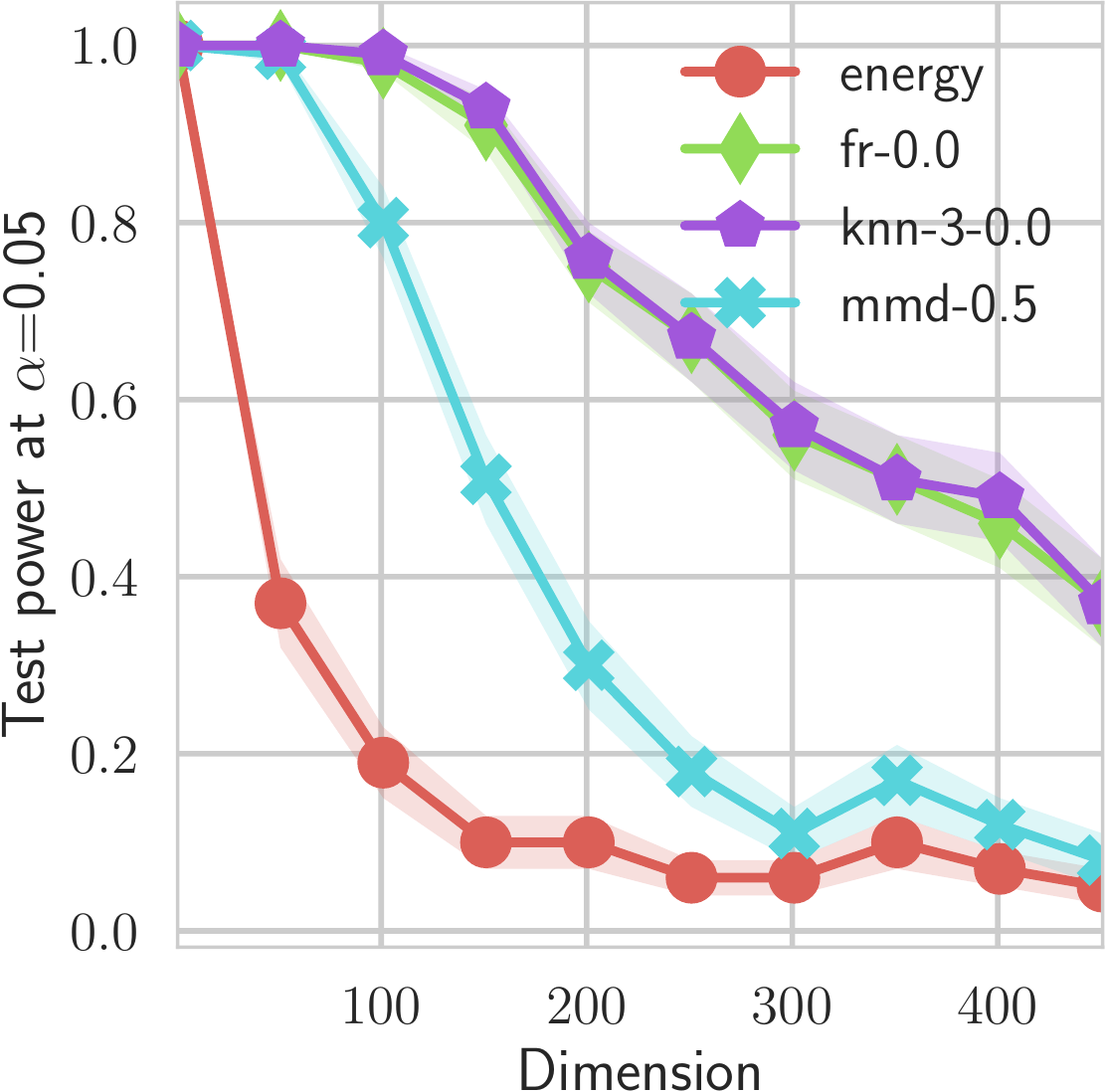}
		}
		\caption{\footnotesize Power against the alternative $(\mu=0, \sigma=3)$ from $n_1=n_2=128$ samples.}

	\end{subfigure}

	\begin{subfigure}[t]{\linewidth}
		{\footnotesize
			\includegraphics[width=0.32\linewidth]{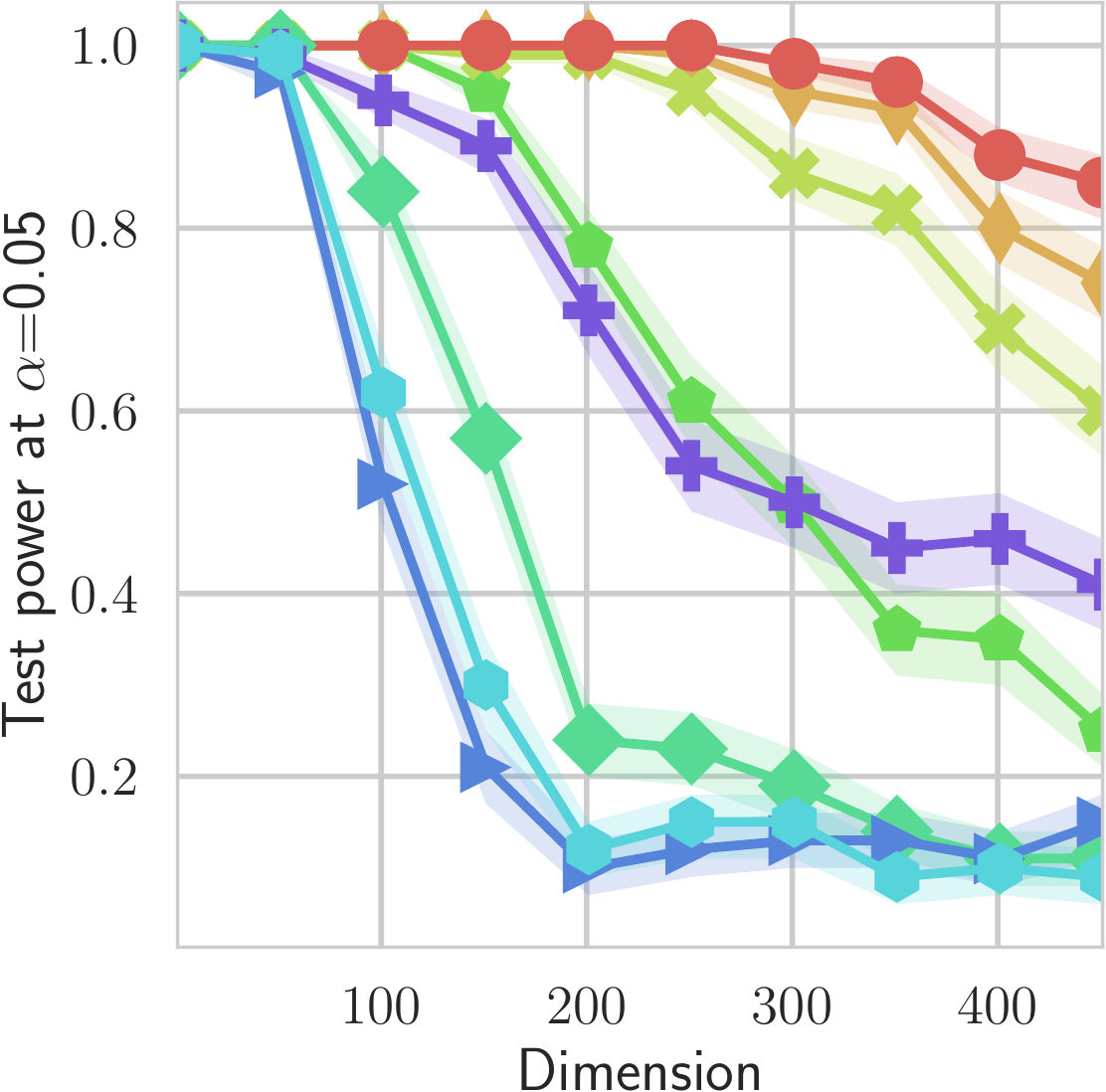}
			\includegraphics[width=0.32\linewidth]{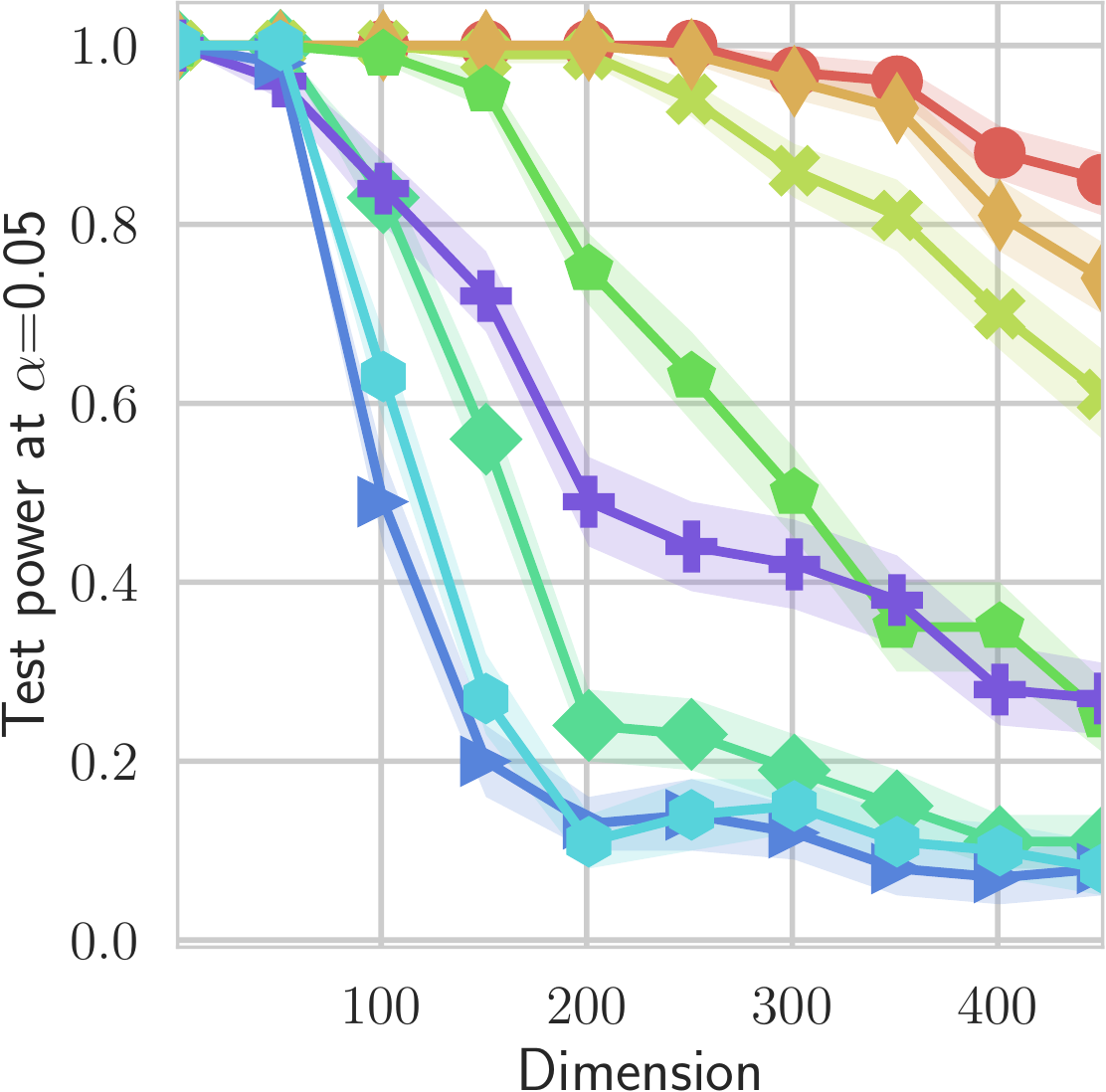}
			\includegraphics[width=0.32\linewidth]{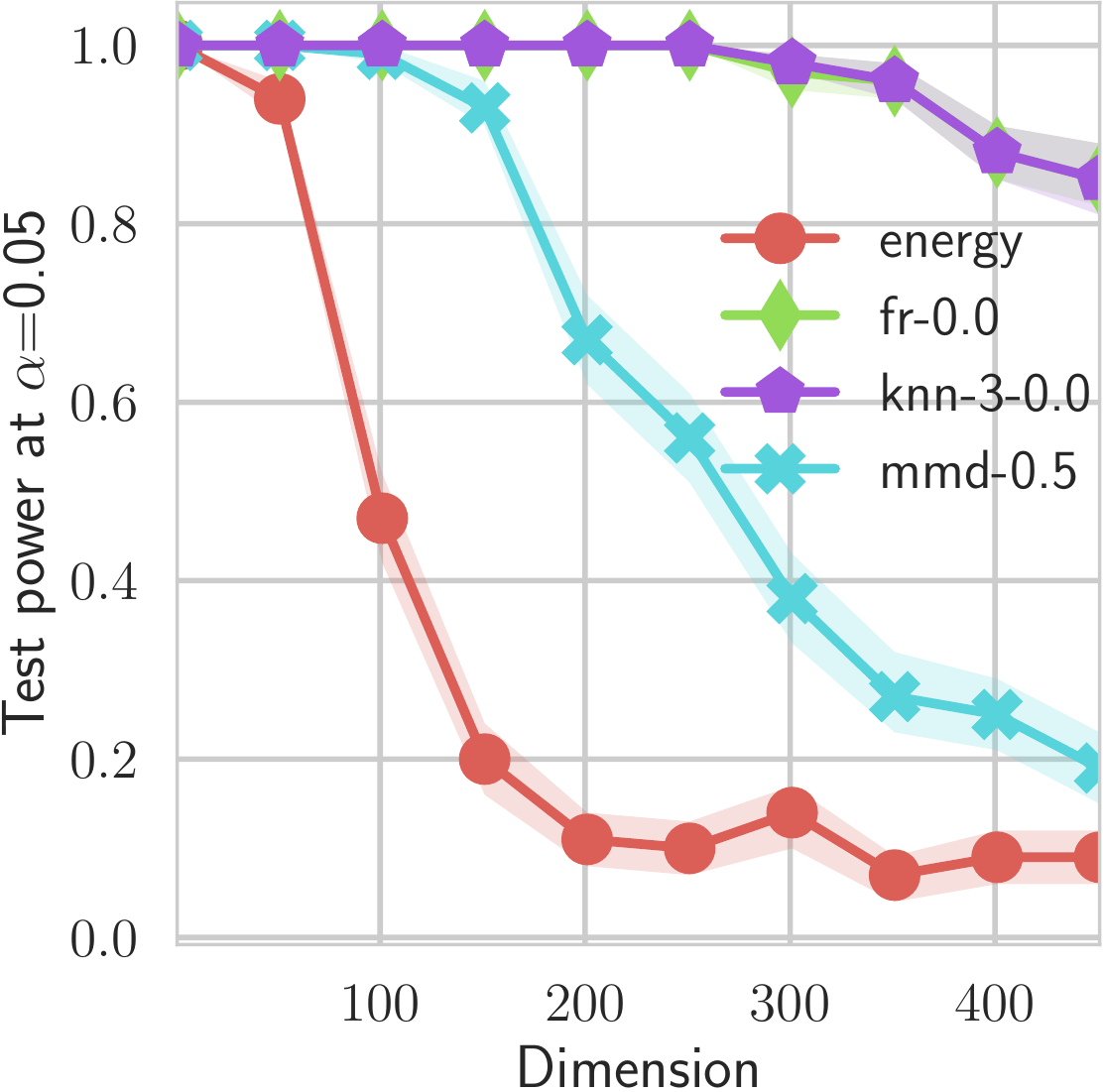}
		}
		\caption{\footnotesize Power against the alternative $(\mu=0, \sigma=3)$ from $n_1=n_2=256$ samples.}
	\end{subfigure}
	\caption{\small Test power when comparing two normal distributions. In the first two columns we present the $3$-NN and FR tests as we vary $\lambda$ ---
		we use fr-$\gamma$ for $\lambda=d^\gamma$, and fr-ct for the classical test (analogously for $3$-NN).
		The legends presented in the first row are consistent across the respective columns.
		The last column compares the best performing of these tests with the best performing MMD tests (the remaining MMD plots are provided in \Cref{app:experiments}).
		Note that our smoothed tests have the largest power, and they significantly outperform their classical counterparts.}\label{fig:power}
\end{figure*}

\paragraph{Power as a function of $\lambda$ and $d$.}
In our first experiment we analyze the effect of the smoothing strength on the power of our differentiable tests.
In addition to the classical FR and $k$-NN tests, we have considered the unbiased MMD test~\cite{gretton2012kernel} with the squared exponential kernel (as implemented in \texttt{Shogun}~\cite{sonnenburg2010shogun} using the code from~\cite{sutherland2016generative}), and the energy test~\cite{szekely2013energy}.
The problem that we consider, which is challenging in high dimensions, is that of differentiating the distribution $\mathcal{N}(\mathbf 0, I)$ from $\mathcal{N}((\mu, 0, \ldots, 0), \texttt{diag}(\sigma^2, 1, \ldots, 1))$.
This setting was considered to be fair in~\cite{ramdas2015decreasing}, as the KL divergence between the distribution is constant irrespective of the dimension.
To set the smoothing strength and the bandwidth of the MMD kernel (in addition to the median heuristic) we used the same strategy as in \cite{ramdas2015decreasing} by setting $\lambda=d^\gamma$ for varying $\gamma\in[0,1]$.
The results are presented in \Cref{fig:power}, where can observe that (i) our test have similar results with MMD for shift-alternatives, while performing significantly better for scale alternatives, and (ii) by varying the smoothing parameter we can significantly increase the power of the test.
In the third column we present only the best performing MMD, while we present the remaining results in \Cref{app:experiments}.
Note that we expect the power to go to zero as the dimension increases~\cite{bhattacharya2015power,ramdas2015decreasing}.

\paragraph{Learning.} As we have already hinted in the introduction, we stochastically optimize
\[ \textrm{max.}_{\btheta}\, \mathbb{E}_{\x_i\sim P, \z_i\sim Q_0}[ t_{\pi^*}^{\lambda}(\{\x_i\}_{i=1}^{n_1}, \{ f_\btheta(\z_i) \}_{i=n_1}^{n_1+n_2})] \]
using the Adam~\cite{kingma2014adam} optimizer.
To optimize, we draw at each round $n_1$ samples from the true distribution $P$, $n_2=n_1$ samples from the base measure $Q_0$, and then plug them in into the smoothed $t$-statistic.

\begin{figure*}
	\centering

	\begin{subfigure}[t]{0.24\linewidth}
		\hfill\includegraphics[width=\textwidth]{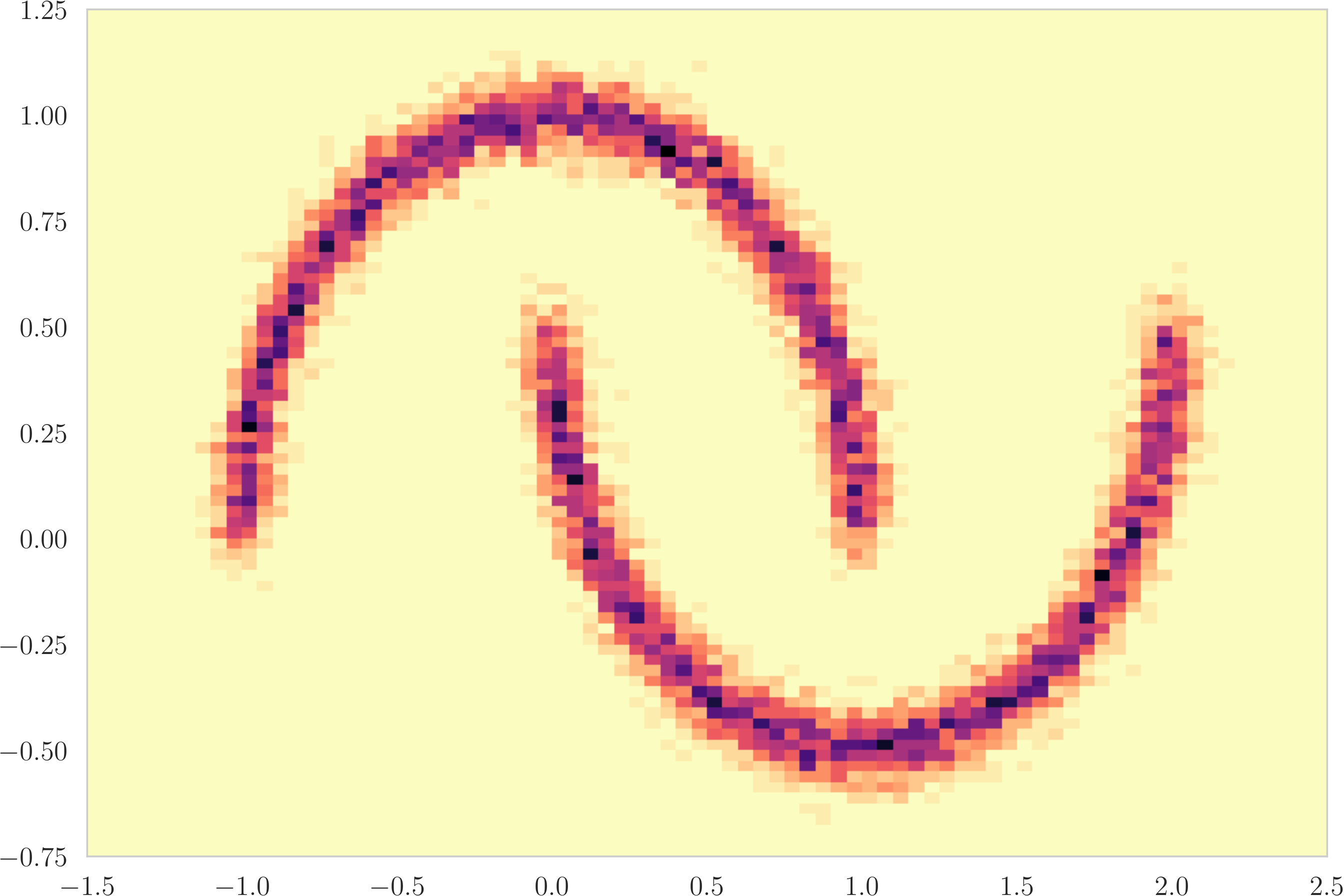}\hspace{0.1cm}
		\caption{Original data.}
	\end{subfigure}
	\begin{subfigure}[t]{0.24\linewidth}
		\centering
		\hfill\includegraphics[width=\textwidth]{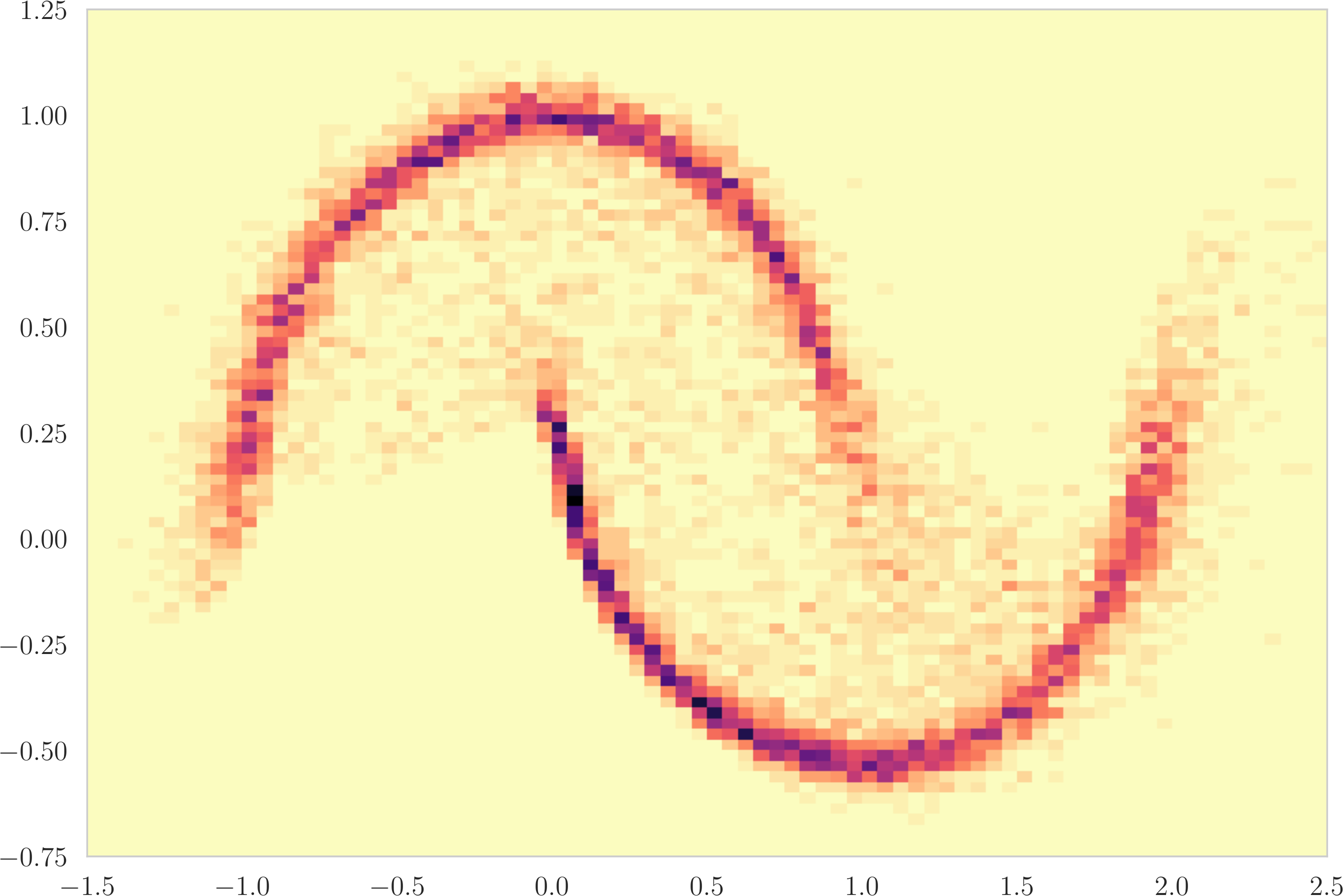}\hspace{0.3cm}
		\includegraphics[width=\textwidth]{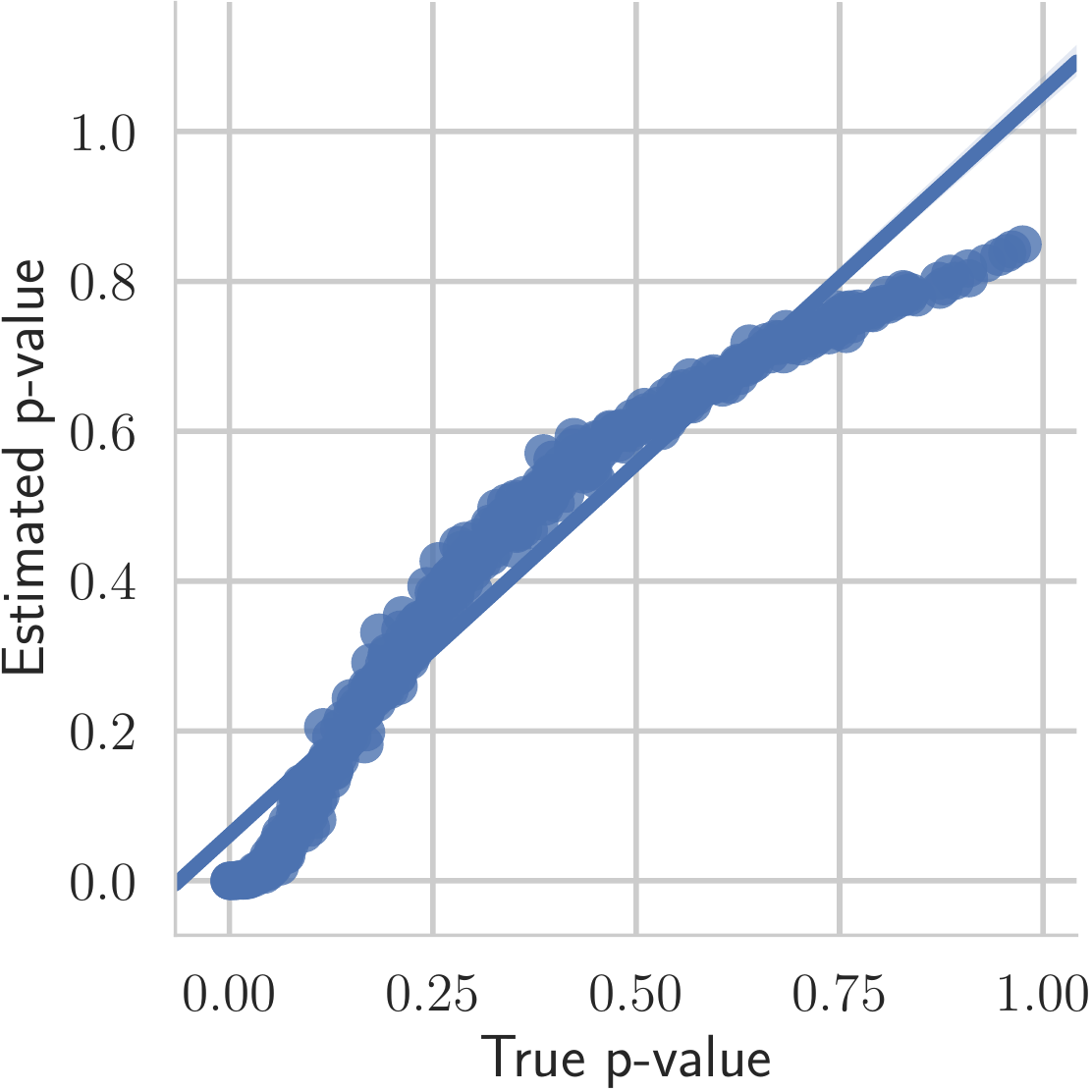}
		\caption{$1$-NN with $\lambda=10$.}\label{fig:moons-a}
	\end{subfigure}
	\begin{subfigure}[t]{0.24\linewidth}
		\centering
		\hfill\includegraphics[width=\textwidth]{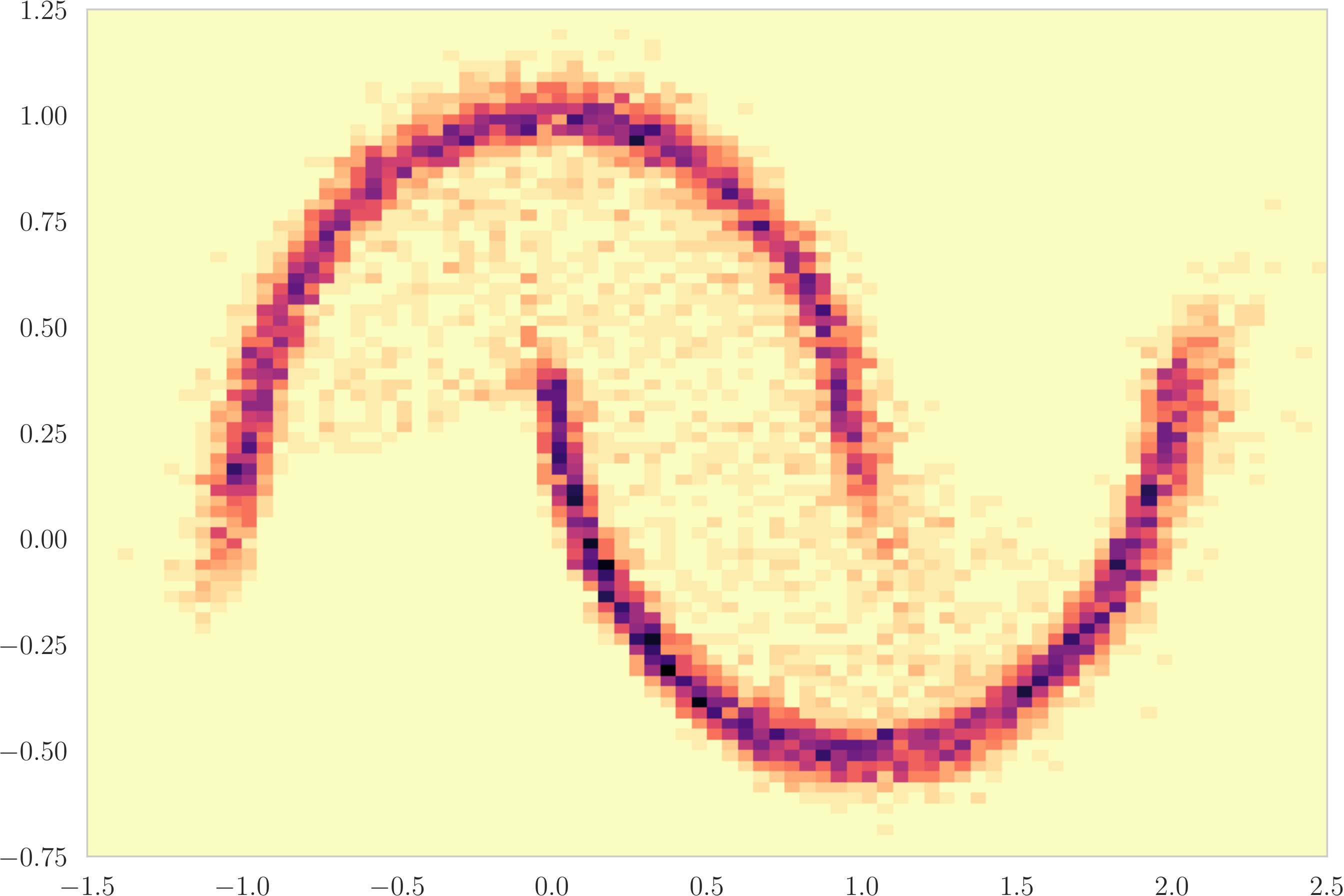}\hspace{0.3cm}
		\includegraphics[width=\textwidth]{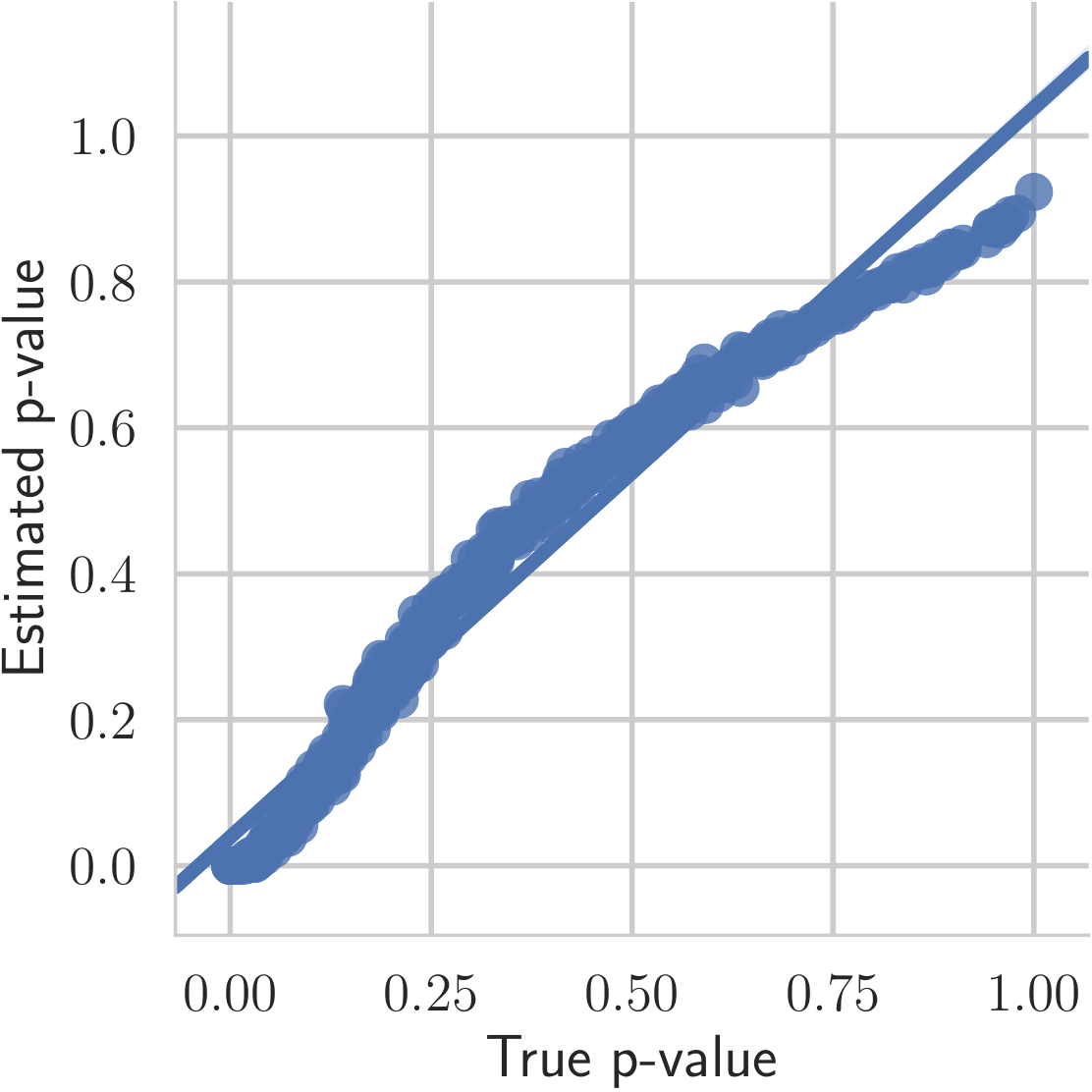}
		\caption{$1$-NN with $\lambda=1$.}\label{fig:moons-b}
	\end{subfigure}
	\begin{subfigure}[t]{0.24\linewidth}
		\hfill\includegraphics[width=\textwidth]{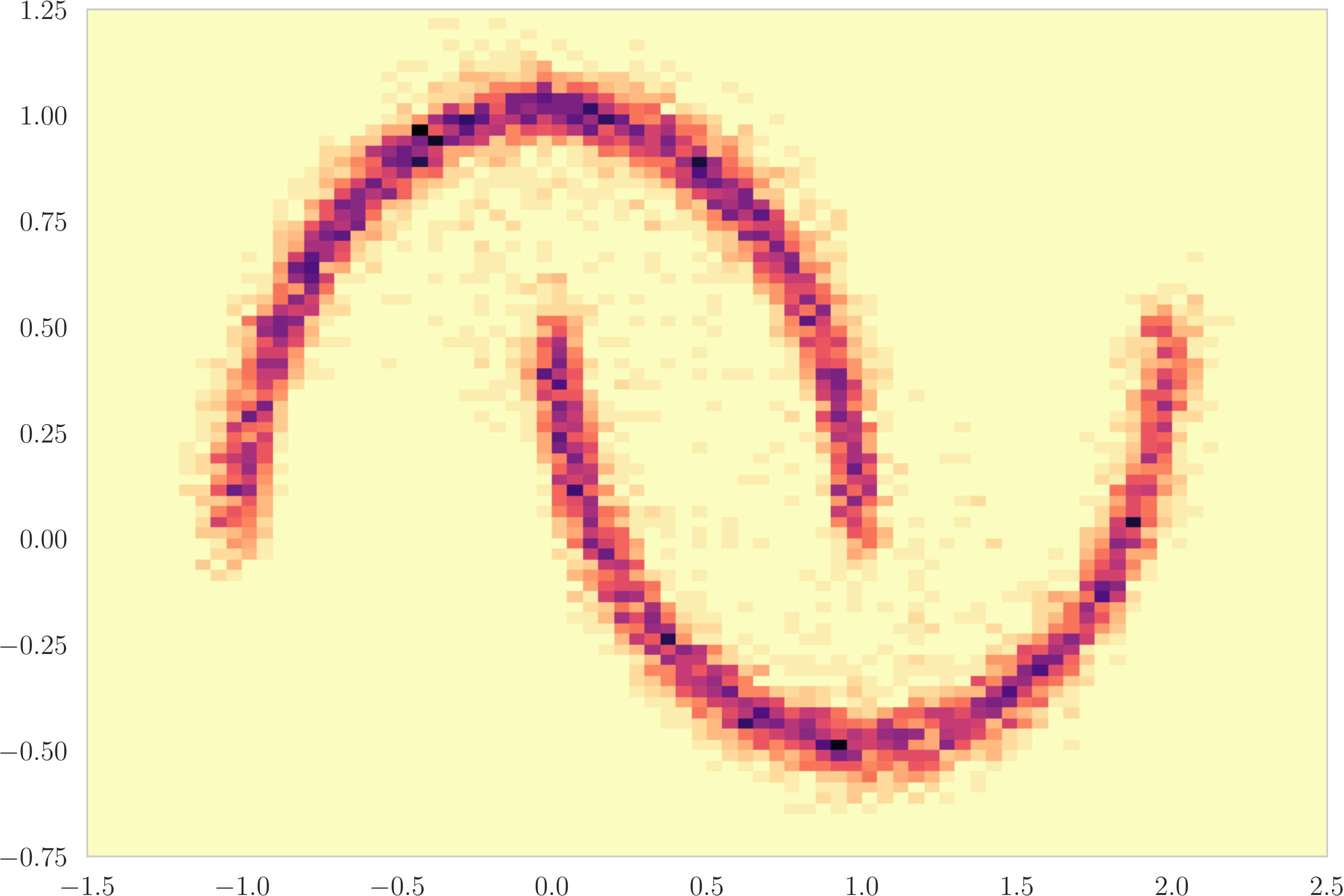}\hspace{0.3cm}
		\includegraphics[width=\textwidth]{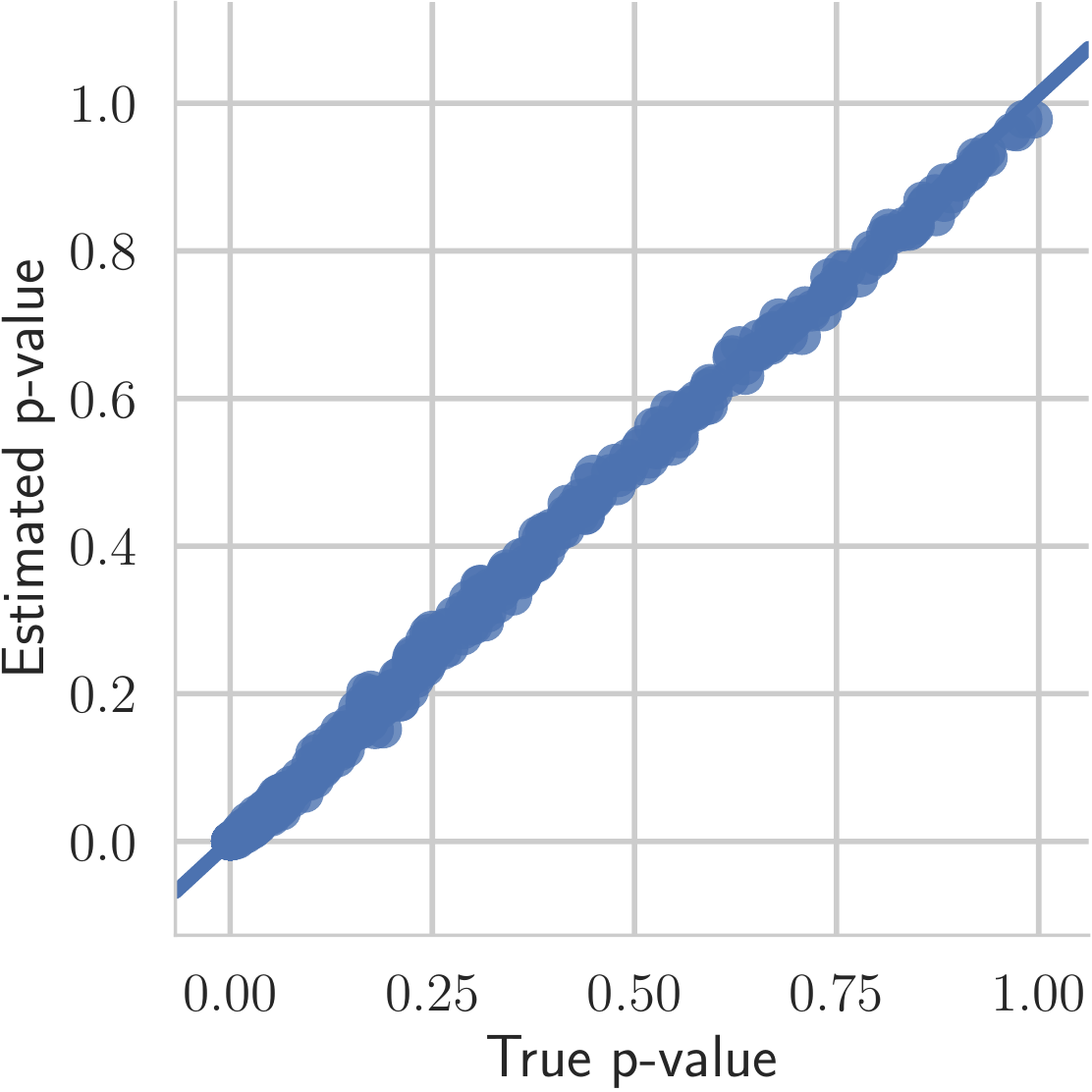}
		\caption{$1$-NN with $\lambda=0.05$.}\label{fig:moons-c}
	\end{subfigure}
	\caption{The effect of varying $\lambda$ on the learned model and the normality of the null statistic. Note that with decreasing $\lambda$ we get closer to normality, and the learned distribution better models the true one.}
	\label{fig:moons}
\end{figure*}

The first experiment we perform, with the goal of understanding the effects of $\lambda$, is on the toy \emph{two moons} dataset from \texttt{scikit-learn}~\cite{scikit-learn}.
We show the results in \Cref{fig:moons}.
From the second row, showing the estimated $p$-value versus the correct one (from 1000 random permutations) at several points during training, we can indeed see that the permutation null gets closer to normality as $\lambda$ decreases.
Most importantly, note that the relationship is monotone, so that we would expect the optimization to not be significantly harmed if we use the approximation.
Qualitatively, we can observe that the solutions have the general structure of $P$, and that they improve as we decrease $\lambda$ --- the symmetry is better captured and the two moons get better separated.

\begin{figure*}
	\begin{subfigure}[t]{0.24\linewidth}
		\includegraphics[width=\linewidth]{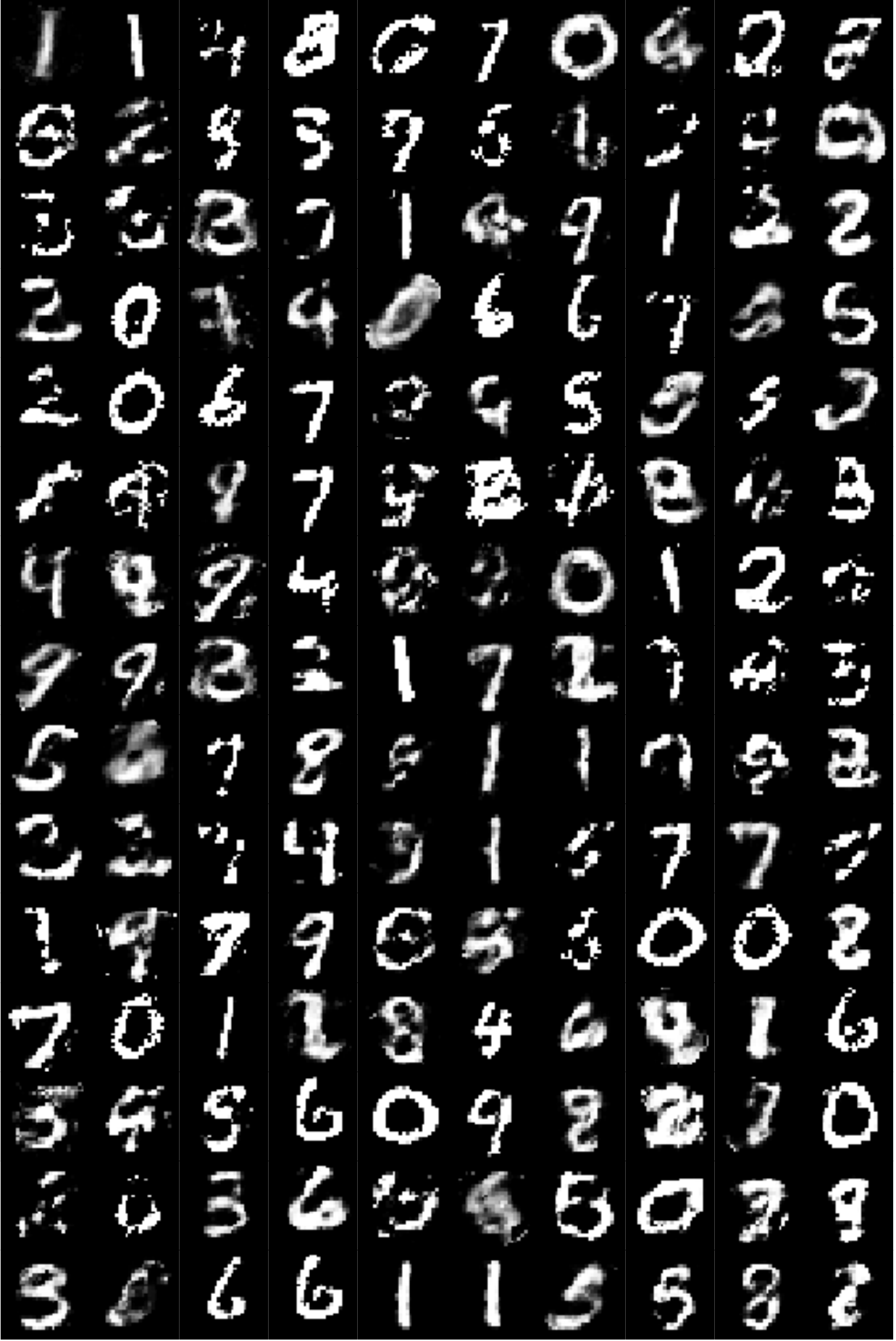}
		\caption{$1$-NN with $\lambda=10$ and $n_1=256$.}\label{fig:mnist-a}
	\end{subfigure}
	\begin{subfigure}[t]{0.24\linewidth}
		\includegraphics[width=\linewidth]{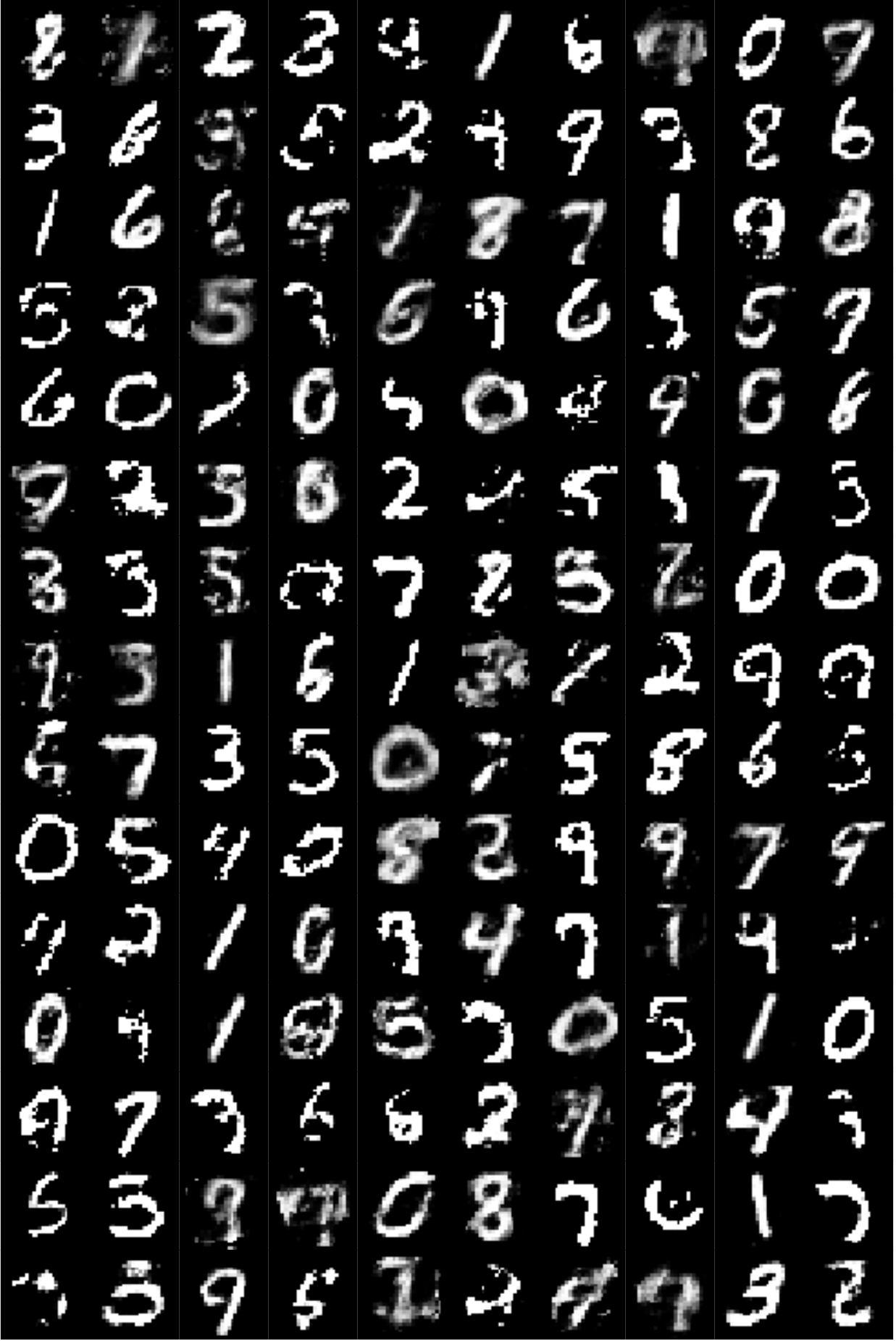}
		\caption{$1$-NN with $\lambda=10$ and $n_1=512$.}\label{fig:mnist-b}
	\end{subfigure}
	\begin{subfigure}[t]{0.24\linewidth}
		\includegraphics[width=\linewidth]{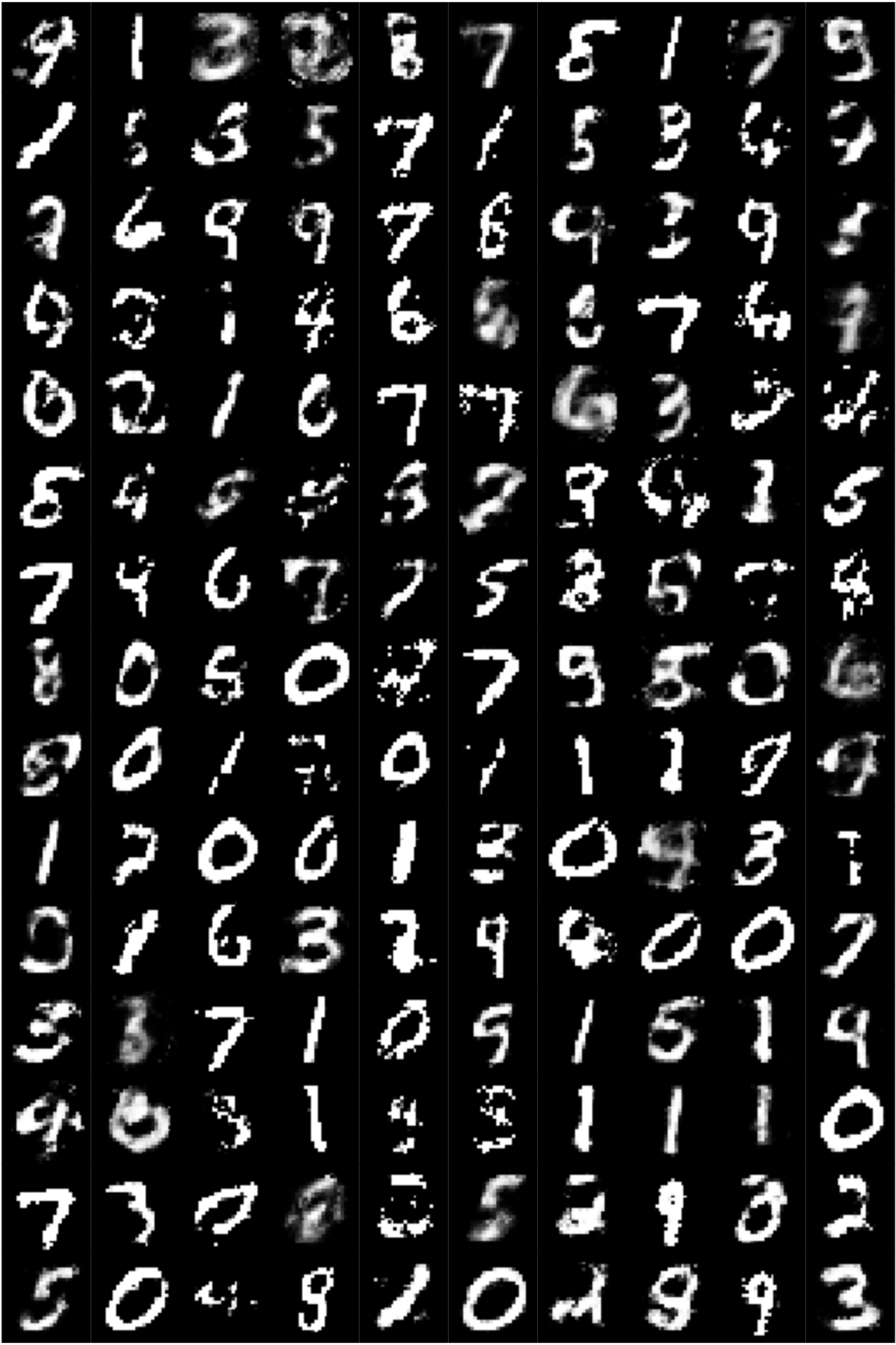}
		\caption{FR $\lambda=10$ and $n_1=128$.}\label{fig:mnist-c}
	\end{subfigure}
	\begin{subfigure}[t]{0.24\linewidth}
		\includegraphics[width=\linewidth]{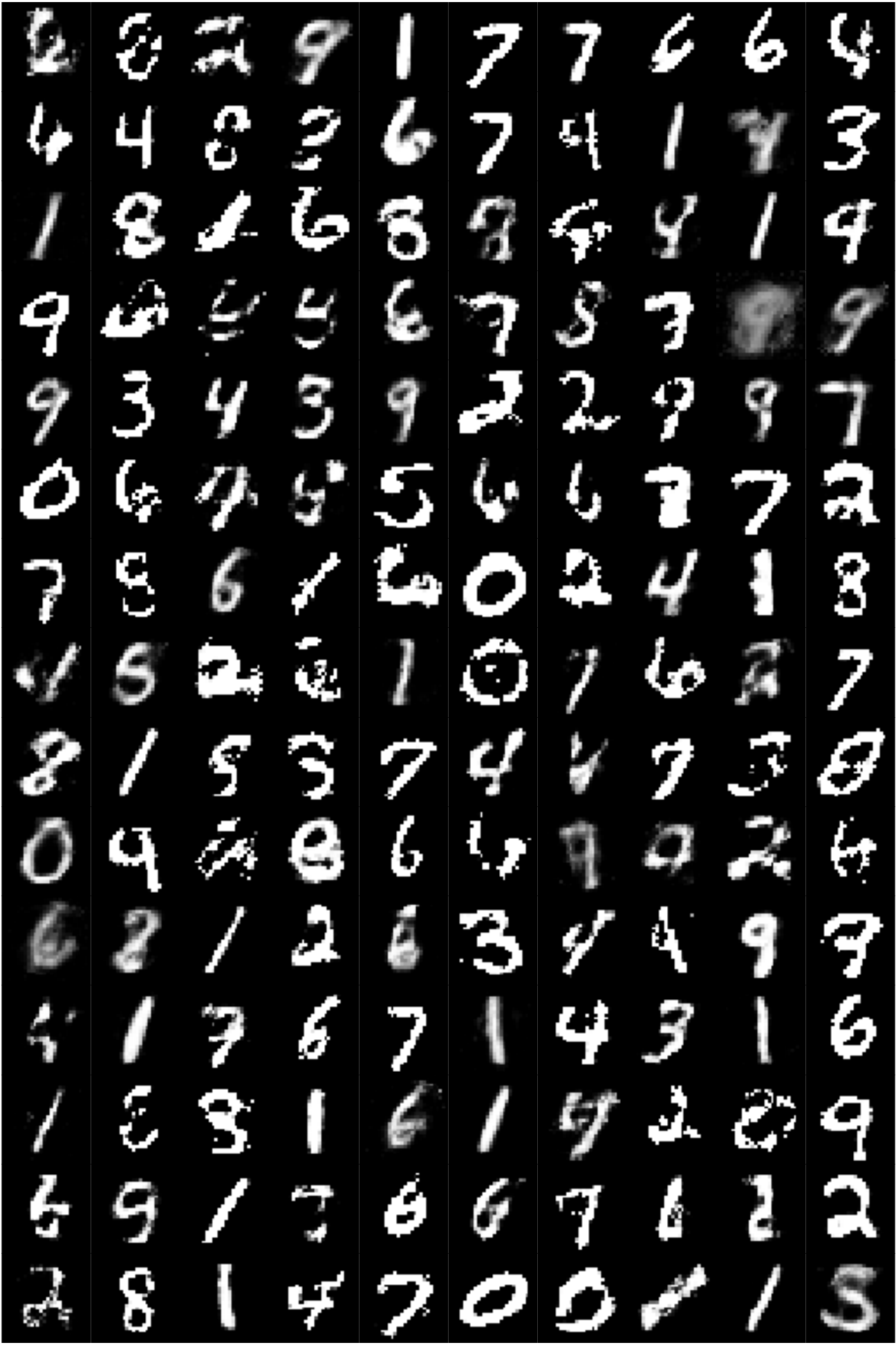}
		\caption{FR $\lambda=5$ and $n_1=128$.}\label{fig:mnist-d}
	\end{subfigure}
	\caption{Four different models trained on MNIST.} \label{fig:mnist}
\end{figure*}

\paragraph{MNIST.} Finally, we have trained several models on the MNIST~\cite{lecun1998gradient} dataset, which we present in \Cref{fig:mnist}.
We can observe that despite the high (784) dimensional data and the fact that we use the distance directly on the pixels, the learned models generate digits that look mostly realistic and are competitive with those obtained using MMD~\cite{li2015generative,dziugaite2015training}.

\section{Conclusion}
We have developed smoothed two-sample graph tests that can be used for learning implicit models.
These tests moreover outperform their classical equivalents on the problem of two sample testing.
We have shown how to compute them by performing inference in undirected models, and presented alternative interpretations by drawing connections to neighbourhood component analysis and spectral graph sparsifiers.
In the last section we have experimentally showcased the benefits of our approach, and presented results from a learned model.

{\footnotesize \paragraph{\footnotesize Acknowledgements.} The research was partially supported by ERC StG 307036 and a Google European PhD Fellowship.}
\bibliography{refs}

\begin{thebibliography}{41}
\providecommand{\natexlab}[1]{#1}
\providecommand{\url}[1]{\texttt{#1}}
\expandafter\ifx\csname urlstyle\endcsname\relax
  \providecommand{\doi}[1]{doi: #1}\else
  \providecommand{\doi}{doi: \begingroup \urlstyle{rm}\Url}\fi

\bibitem[Wald and Wolfowitz(1940)]{wald1940test}
Abraham Wald and Jacob Wolfowitz.
\newblock On a test whether two samples are from the same population.
\newblock \emph{Annals of Mathematical Statistics}, 11\penalty0 (2):\penalty0
  147--162, 1940.

\bibitem[Friedman and Rafsky(1979)]{friedman1979multivariate}
Jerome~H Friedman and Lawrence~C Rafsky.
\newblock Multivariate generalizations of the wald-wolfowitz and smirnov
  two-sample tests.
\newblock \emph{Annals of Statistics}, pages 697--717, 1979.

\bibitem[Friedman and Rafsky(1983)]{friedman1983graph}
Jerome~H Friedman and Lawrence~C Rafsky.
\newblock Graph-theoretic measures of multivariate association and prediction.
\newblock \emph{Annals of Statistics}, pages 377--391, 1983.

\bibitem[Henze and Penrose(1999)]{henze1999multivariate}
Norbert Henze and Mathew~D Penrose.
\newblock On the multivariate runs test.
\newblock \emph{Annals of Statistics}, pages 290--298, 1999.

\bibitem[Henze(1988)]{henze1988multivariate}
Norbert Henze.
\newblock A multivariate two-sample test based on the number of nearest
  neighbor type coincidences.
\newblock \emph{Annals of Statistics}, pages 772--783, 1988.

\bibitem[Schilling(1986)]{schilling1986multivariate}
Mark~F Schilling.
\newblock Multivariate two-sample tests based on nearest neighbors.
\newblock \emph{Journal of the American Statistical Association}, 81\penalty0
  (395):\penalty0 799--806, 1986.

\bibitem[Bhattacharya(2015)]{bhattacharya2015power}
Bhaswar~B Bhattacharya.
\newblock Power of graph-based two-sample tests.
\newblock \emph{arXiv preprint arXiv:1508.07530}, 2015.

\bibitem[Chen and Zhang(2013)]{chen2013graph}
Hao Chen and Nancy~R Zhang.
\newblock Graph-based tests for two-sample comparisons of categorical data.
\newblock \emph{Statistica Sinica}, pages 1479--1503, 2013.

\bibitem[Gretton et~al.(2012)Gretton, Borgwardt, Rasch, Sch{\"o}lkopf, and
  Smola]{gretton2012kernel}
Arthur Gretton, Karsten~M Borgwardt, Malte~J Rasch, Bernhard Sch{\"o}lkopf, and
  Alexander Smola.
\newblock A kernel two-sample test.
\newblock \emph{Journal of Machine Learning Research}, 13\penalty0
  (Mar):\penalty0 723--773, 2012.

\bibitem[Li et~al.(2015)Li, Swersky, and Zemel]{li2015generative}
Yujia Li, Kevin Swersky, and Rich Zemel.
\newblock Generative moment matching networks.
\newblock In \emph{International Conference on Machine Learning (ICML)}, 2015.

\bibitem[Dziugaite et~al.(2015)Dziugaite, Roy, and
  Ghahramani]{dziugaite2015training}
Gintare~Karolina Dziugaite, Daniel~M. Roy, and Zoubin Ghahramani.
\newblock Training generative neural networks via maximum mean discrepancy
  optimization.
\newblock In \emph{Uncertainty in Artificial Intelligence (UAI)}, 2015.

\bibitem[Sutherland et~al.(2016)Sutherland, Tung, Strathmann, De, Ramdas,
  Smola, and Gretton]{sutherland2016generative}
Dougal~J Sutherland, Hsiao-Yu Tung, Heiko Strathmann, Soumyajit De, Aaditya
  Ramdas, Alex Smola, and Arthur Gretton.
\newblock Generative models and model criticism via optimized maximum mean
  discrepancy.
\newblock In \emph{International Conference on Learning Representations
  (ICLR)}, 2016.

\bibitem[Sz{\'e}kely and Rizzo(2013)]{szekely2013energy}
G{\'a}bor~J Sz{\'e}kely and Maria~L Rizzo.
\newblock Energy statistics: A class of statistics based on distances.
\newblock \emph{Journal of Statistical Planning and Inference}, 143\penalty0
  (8):\penalty0 1249--1272, 2013.

\bibitem[Bellemare et~al.(2017)Bellemare, Danihelka, Dabney, Mohamed,
  Lakshminarayanan, Hoyer, and Munos]{bellemare2017cramer}
Marc~G Bellemare, Ivo Danihelka, Will Dabney, Shakir Mohamed, Balaji
  Lakshminarayanan, Stephan Hoyer, and R{\'e}mi Munos.
\newblock The cramer distance as a solution to biased wasserstein gradients.
\newblock \emph{arXiv preprint arXiv:1705.10743}, 2017.

\bibitem[Sugiyama et~al.(2012)Sugiyama, Suzuki, and
  Kanamori]{sugiyama2012density}
Masashi Sugiyama, Taiji Suzuki, and Takafumi Kanamori.
\newblock \emph{Density ratio estimation in machine learning}.
\newblock Cambridge University Press, 2012.

\bibitem[Goodfellow et~al.(2014)Goodfellow, Pouget-Abadie, Mirza, Xu,
  Warde-Farley, Ozair, Courville, and Bengio]{goodfellow2014generative}
Ian Goodfellow, Jean Pouget-Abadie, Mehdi Mirza, Bing Xu, David Warde-Farley,
  Sherjil Ozair, Aaron Courville, and Yoshua Bengio.
\newblock Generative adversarial nets.
\newblock In \emph{Advances in Neural Information Processing Systems (NIPS)},
  pages 2672--2680, 2014.

\bibitem[Salimans et~al.(2016)Salimans, Goodfellow, Zaremba, Cheung, Radford,
  and Chen]{salimans2016improved}
Tim Salimans, Ian Goodfellow, Wojciech Zaremba, Vicki Cheung, Alec Radford, and
  Xi~Chen.
\newblock Improved techniques for training gans.
\newblock In \emph{Advances in Neural Information Processing Systems (NIPS)},
  pages 2234--2242, 2016.

\bibitem[Nowozin et~al.(2016)Nowozin, Cseke, and Tomioka]{nowozin2016f}
Sebastian Nowozin, Botond Cseke, and Ryota Tomioka.
\newblock $f$-{GAN}: Training generative neural samplers using variational
  divergence minimization.
\newblock In \emph{Advances in Neural Information Processing Systems (NIPS)},
  pages 271--279, 2016.

\bibitem[Ali and Silvey(1966)]{ali1966general}
Syed~Mumtaz Ali and Samuel~D Silvey.
\newblock A general class of coefficients of divergence of one distribution
  from another.
\newblock \emph{Journal of the Royal Statistical Society. Series B
  (Methodological)}, pages 131--142, 1966.

\bibitem[Mohamed and Lakshminarayanan(2016)]{mohamed2016learning}
Shakir Mohamed and Balaji Lakshminarayanan.
\newblock Learning in implicit generative models.
\newblock \emph{arXiv preprint arXiv:1610.03483}, 2016.

\bibitem[Prim(1957)]{prim1957shortest}
Robert~Clay Prim.
\newblock Shortest connection networks and some generalizations.
\newblock \emph{Bell Labs Technical Journal}, 36\penalty0 (6):\penalty0
  1389--1401, 1957.

\bibitem[Kruskal(1956)]{kruskal1956shortest}
Joseph~B Kruskal.
\newblock On the shortest spanning subtree of a graph and the traveling
  salesman problem.
\newblock \emph{Proceedings of the American Mathematical society}, 7\penalty0
  (1):\penalty0 48--50, 1956.

\bibitem[Berisha and Hero(2015)]{berisha2015empirical}
Visar Berisha and Alfred~O Hero.
\newblock Empirical non-parametric estimation of the fisher information.
\newblock \emph{IEEE Signal Processing Letters}, 22\penalty0 (7):\penalty0
  988--992, 2015.

\bibitem[Wainwright and Jordan(2008)]{wainwright2008graphical}
Martin~J Wainwright and Michael~I Jordan.
\newblock Graphical models, exponential families, and variational inference.
\newblock \emph{Foundations and Trends{\textregistered} in Machine Learning},
  1\penalty0 (1-2), 2008.

\bibitem[Daniels(1944)]{daniels1944relation}
Henry~E Daniels.
\newblock The relation between measures of correlation in the universe of
  sample permutations.
\newblock \emph{Biometrika}, 33\penalty0 (2):\penalty0 129--135, 1944.

\bibitem[Barbour and Eagleson(1986)]{barbour1986random}
AD~Barbour and GK~Eagleson.
\newblock Random association of symmetric arrays.
\newblock \emph{Stochastic Analysis and Applications}, 4\penalty0 (3):\penalty0
  239--281, 1986.

\bibitem[Yukich(2006)]{yukich2006probability}
Joseph~E Yukich.
\newblock \emph{Probability theory of classical Euclidean optimization
  problems}.
\newblock Springer, 2006.

\bibitem[Tarlow et~al.(2012)Tarlow, Swersky, Zemel, Adams, and
  Frey]{tarlow2012fast}
Daniel Tarlow, Kevin Swersky, Richard~S Zemel, Ryan~Prescott Adams, and
  Brendan~J Frey.
\newblock Fast exact inference for recursive cardinality models.
\newblock \emph{Uncertainty in Artificial Intelligence (UAI)}, 2012.

\bibitem[Swersky et~al.(2012)Swersky, Sutskever, Tarlow, Zemel, Salakhutdinov,
  and Adams]{swersky2012cardinality}
Kevin Swersky, Ilya Sutskever, Daniel Tarlow, Richard~S Zemel, Ruslan~R
  Salakhutdinov, and Ryan~P Adams.
\newblock Cardinality restricted boltzmann machines.
\newblock In \emph{Advances in Neural Information Processing Systems (NIPS)},
  pages 3293--3301, 2012.

\bibitem[Goldberger et~al.(2005)Goldberger, Hinton, Roweis, and
  Salakhutdinov]{goldberger2005neighbourhood}
Jacob Goldberger, Geoffrey~E Hinton, Sam~T Roweis, and Ruslan~R Salakhutdinov.
\newblock Neighbourhood components analysis.
\newblock In \emph{Advances in Neural Information Processing Systems (NIPS)},
  pages 513--520, 2005.

\bibitem[Tarlow et~al.(2013)Tarlow, Swersky, Charlin, Sutskever, and
  Zemel]{tarlow2013stochastic}
Daniel Tarlow, Kevin Swersky, Laurent Charlin, Ilya Sutskever, and Rich Zemel.
\newblock Stochastic k-neighborhood selection for supervised and unsupervised
  learning.
\newblock In \emph{International Conference on Machine Learning}, pages
  199--207, 2013.

\bibitem[Lyons(2003)]{lyons2003determinantal}
Russell Lyons.
\newblock Determinantal probability measures.
\newblock \emph{Publications math{\'e}matiques de l'IH{\'E}S}, 98\penalty0
  (1):\penalty0 167--212, 2003.

\bibitem[Chandra et~al.(1996)Chandra, Raghavan, Ruzzo, Smolensky, and
  Tiwari]{chandra1996electrical}
Ashok~K Chandra, Prabhakar Raghavan, Walter~L Ruzzo, Roman Smolensky, and
  Prasoon Tiwari.
\newblock The electrical resistance of a graph captures its commute and cover
  times.
\newblock \emph{Computational Complexity}, 6\penalty0 (4):\penalty0 312--340,
  1996.

\bibitem[Spielman and Srivastava(2011)]{spielman2011graph}
Daniel~A Spielman and Nikhil Srivastava.
\newblock Graph sparsification by effective resistances.
\newblock \emph{SIAM Journal on Computing}, 40\penalty0 (6):\penalty0
  1913--1926, 2011.

\bibitem[Spielman and Teng(2014)]{spielman2014nearly}
Daniel~A Spielman and Shang-Hua Teng.
\newblock Nearly linear time algorithms for preconditioning and solving
  symmetric, diagonally dominant linear systems.
\newblock \emph{SIAM Journal on Matrix Analysis and Applications}, 35\penalty0
  (3):\penalty0 835--885, 2014.

\bibitem[Spielman and Teng(2011)]{spielman2011spectral}
Daniel~A Spielman and Shang-Hua Teng.
\newblock Spectral sparsification of graphs.
\newblock \emph{SIAM Journal on Computing}, 40\penalty0 (4):\penalty0
  981--1025, 2011.

\bibitem[Sonnenburg et~al.(2010)Sonnenburg, Henschel, Widmer, Behr, Zien, Bona,
  Binder, Gehl, Franc, et~al.]{sonnenburg2010shogun}
S{\'C}~Sonnenburg, Sebastian Henschel, Christian Widmer, Jonas Behr, Alexander
  Zien, Fabio~de Bona, Alexander Binder, Christian Gehl, Vojt{\"A} Franc,
  et~al.
\newblock The shogun machine learning toolbox.
\newblock \emph{Journal of Machine Learning Research}, 11\penalty0
  (Jun):\penalty0 1799--1802, 2010.

\bibitem[Ramdas et~al.(2015)Ramdas, Reddi, P{\'o}czos, Singh, and
  Wasserman]{ramdas2015decreasing}
Aaditya Ramdas, Sashank~Jakkam Reddi, Barnab{\'a}s P{\'o}czos, Aarti Singh, and
  Larry~A Wasserman.
\newblock On the decreasing power of kernel and distance based nonparametric
  hypothesis tests in high dimensions.
\newblock In \emph{AAAI}, 2015.

\bibitem[Kingma and Ba(2015)]{kingma2014adam}
Diederik Kingma and Jimmy Ba.
\newblock Adam: A method for stochastic optimization.
\newblock In \emph{International Conference on Learning Representations
  (ICLR)}, 2015.

\bibitem[Pedregosa et~al.(2011)Pedregosa, Varoquaux, Gramfort, Michel, Thirion,
  Grisel, Blondel, Prettenhofer, Weiss, Dubourg, Vanderplas, Passos,
  Cournapeau, Brucher, Perrot, and Duchesnay]{scikit-learn}
F.~Pedregosa, G.~Varoquaux, A.~Gramfort, V.~Michel, B.~Thirion, O.~Grisel,
  M.~Blondel, P.~Prettenhofer, R.~Weiss, V.~Dubourg, J.~Vanderplas, A.~Passos,
  D.~Cournapeau, M.~Brucher, M.~Perrot, and E.~Duchesnay.
\newblock Scikit-learn: Machine learning in {P}ython.
\newblock \emph{Journal of Machine Learning Research}, 12:\penalty0 2825--2830,
  2011.

\bibitem[LeCun et~al.(1998)LeCun, Bottou, Bengio, and
  Haffner]{lecun1998gradient}
Yann LeCun, L{\'e}on Bottou, Yoshua Bengio, and Patrick Haffner.
\newblock Gradient-based learning applied to document recognition.
\newblock \emph{Proceedings of the IEEE}, 86\penalty0 (11):\penalty0
  2278--2324, 1998.

\end{thebibliography}
\newpage
\onecolumn
\appendix
\section{Proofs}

\begin{proof}[Proof of \cref{thm:moments}]
	The expectation of the statistic under $H_0$ is (when $\pi$ is a uniformly random labelling)
	\[
		\sum_{e\in E} \mu(\w/\lambda)_e \underbrace{\E_\pi[\Delta_{\pi}(e)]}_{2 n_1n_2/n(n-1)} = 2 m n_1n_2/n(n-1),
	\]
	where the inner expectation $\E_\pi[\Delta_{\pi}(e)]$ has been computed in~\cite{friedman1979multivariate}.
	We can also easily compute the variance as
	\begin{align}\label{eqn:app-var}
		\sum_{e,e'\in E} \CoVar_{\pi\sim H_0}[\mu_e\Delta_\pi(e), \mu_{e'}\Delta_\pi(e')] & = \sum_{e,e'\in E} \mu_e\mu_{e'}\underbrace{\E_{\pi\sim H_0}[\Delta_\pi(e) \Delta_\pi(e')]}_{\Pi_{e,e'}} - \underbrace{\frac{4n_1^2n_2^2}{n^2(n-1)^2} m^2}_{(\E_{\pi\sim H_0}[T^\lambda_{\pi^*}])^2}.
	\end{align}
\end{proof}

\begin{proof}[Proof of \cref{lem:simple-var}]
	We can split the sum in the variance formula over all edge pairs into three groups as follows
	\begin{equation}\label{eqn:var-split}
		\sum_{e}\sum_{e'\sim e} \underbrace{\frac{n_1n_2}{n(n-1)}}_{\chi_1}\mu_e\mu_{e'} +
		\sum_{e}\sum_{e'\perp e} \underbrace{\frac{4n_1n_2(n_1-1)(n_2-1)}{n(n-1)(n-2)(n-3)}}_{\chi_1\chi_2}\mu_e\mu_{e'} + \sum_e \underbrace{\frac{n_1n_2}{n(n-1)}}_{\chi_1}(\mu_e^2 + \mu_e\mu_{\overline{e}}),
	\end{equation}
	where $\sum_{e'\sim e}$ sums over all edges $e'$ that share at least one vertex with $e$, and $\sum_{e'\perp e}$ sums over those edges that share no vertex with $e$, and $\overline{e}$ denote the \emph{reverse} edge of $e$ (if it exist, zero otherwise).
	Note that each term $\mu_e\mu_{e'}$ appears twice if $e\neq e'$, as in the formula for the variance~\eqref{eqn:app-var}.
	Moreover, note that if $\delta(e)=\delta(e')$, then in the above formula the term $\mu_e\mu_{e'}$ (same for $\mu_{e'}\mu_e$) gets multiplied by $2\chi_1=\Pi_{e,e'}$, as it appears in both the first and the third term.
	Given that assumption that $|U|=m$ under $\nu(\cdot)$, we also know that
	\[
		m^2 = (\sum_e\mu_e)^2 = \sum_e\sum_{e'}\mu_e\mu_{e'} = \sum_e \sum_{e'\sim e} \mu_e\mu_{e'} + \sum_e \sum_{e'\perp e} \mu_e\mu_{e'},
	\]
	so that eq.~\eqref{eqn:var-split} can be simplified to
	\[
		\chi_1 \sum_e \sum_{e'\sim e} \mu_e\mu_{e'} +
		\chi_1\chi_2 (m^2 - \sum_e \sum_{e'\sim e} \mu_e\mu_{e'}) + \chi_1 \sum_e (\mu_e^2 + \mu_e\mu_{\overline{e}}),
	\]
	which be simplified to
	\[
		\chi_1(1-\chi_2) \sum_e \sum_{e'\sim e} \mu_e\mu_{e'} + \chi_1 \sum_e (\mu_e^2 + \mu_e\mu_{\overline{e}}) + \chi_1\chi_2 m^2.
	\]
	Now the result follows by observing that
	\[
		\sum_v (\sum_{e\in \delta(v)}\mu_e)^2 = \sum_{e}\sum_{e'\sim e} \mu_e\mu_{e'} + \sum_e \mu_e^2 + \sum_e \mu_e\mu_{\overline{e}}.
	\]
	To understand why this holds, let us count how many times each term $\mu_e\mu_{e'}$ appears on both sides of the equality if we expand the lhs.
	If $e\neq e'$ and they share exactly one vertex, then the lhs will have two $\mu_e\mu_{e'}$ terms, as $\mu_e$ and $\mu_{e'}$ will be multiplied only at the term corresponding to the shared vertex.
	On the other hand, if $e=e'$ we will again have two $\mu_e\mu_{e'}=\mu_e^2$ terms, as we get one contribution from each end-point of $e$.
	Finally, if $e'=\overline{e}$, we have a total of four $\mu_e\mu_{e'}$ terms, as we get two $\mu_e\mu_{e'}$ from each end-point.
	Thus, eq.~\eqref{eqn:var-split} is equal to
	\[
		\chi_1(1-\chi_2)\big(\sum_v (\sum_{e\in \delta(v)}\mu_e)^2 - \sum_e \mu_e^2 - \sum_e \mu_e\mu_{\overline{e}}\big) + \chi_1 \sum_e (\mu_e^2 + \mu_e\mu_{\overline{e}}) + \chi_1\chi_2 m^2.
	\]
	Finally, if we subtract $4\chi_1^2m^2$ and simplify the expression we have
	\[
		\chi_1(1-\chi_2)\sum_v (\sum_{e\in \delta(v)}\mu_e)^2 + \chi_1\chi_2\sum_e \mu_e^2 + \chi_1\chi_2\sum_e \mu_e\mu_{\overline{e}} + \chi_1(\chi_2 - 4\chi_1) m^2,
	\]
	which is exactly what is claimed in the theorem, if we observe that $e$ and $\overline{e}$ are the only edges parallel to $e$.
\end{proof}

\paragraph{Proof that $\chi_2-4\chi_1\geq 0$ when $n_1=n_2=n/2$.}
First, note that $\frac{n_1}{n} \leq \frac{n_1-1}{n-2}$, if and only if $n_1n-2n_1 \leq nn_1 -n$, which is equivalent to $n_1\geq \frac{1}{2}n$. Similarly, we have $\frac{n_2}{n-1}\leq\frac{n_2-1}{n-3}$ iff $nn_2-3n_2\leq nn_2-n-n_2+1$, which can be re-written as $-2n_2\leq -n+1$, i.e., $n_2\geq\frac{n}{2}-\frac{1}{2}$.
Combining these two inequalities proves the result.

\paragraph{Proof of \Cref{thm:normality}.}
Let us compute an upper bound on the quantities in \cite{barbour1986random}.
\begin{align*}
	a_1 & = \frac{1}{n(n-1)} \sum_{i,j}\omu_{i,j} = \frac{k}{n}                                        & b_1 & = \frac{2}{n(n-1)} n_2n_1 = \Theta(1) \\
	a_2 & = \frac{1}{n(n-1)(n-2)} \underbrace{\sum_{i,j,k}\omu_{i,j}\omu_{i,k}}_{S_2}                  &
	b_2 & = \frac{n_2n_1^2+n_1n_2^2}{n(n-1)(n-2)}=\Theta(1)                                                                                          \\
	a_3 & = \frac{1}{n(n-1)(n-2)(n-3)} \underbrace{\sum_{i,j,k,m}\omu_{i,j}\omu_{i,k}\omu_{i,m}}_{S_3} &
	b_3 & = \frac{n_2n_1^3 + n_1n_2^3}{n(n-1)(n-2)(n-3)}=\Theta(1)                                                                                   \\
	a_4 & = \frac{1}{n(n-1)(n-2)(n-3)} \underbrace{\sum_{i,j,k,m}\omu_{k,i}\omu_{i,j}\omu_{j,m}}_{L_4} &
	b_4 & = 2\frac{n_2^2n_1^2}{n(n-1)(n-2)(n-3)} = \Theta(1)                                                                                              \\
	a_5 & = \frac{1}{n(n-1)(n-2)} \sum_{i,j,k}\omu_{i,j}^2\omu_{i,k} = O(a_2)                          &
	b_5 & = b_2                                                                                                                                      \\
	a_6 & = \frac{1}{n(n-1)} \sum_{i,j}\omu_{i,j}^3 = O(a_1)                                           &
	b_6 & = b_1                                                                                                                                      \\
	a_7 & = \frac{1}{n(n-1)(n-2)} \sum_{i,j,k,m}\omu_{i,j}\omu_{i,k}\omu_{j,k}                         &
	b_7 & = \frac{n_2n_1n_2 + n_1n_2n_1}{n(n-1)(n-2)} = \Theta(1)                                                                                    \\
	a_8 & = \frac{1}{n(n-1)} \sum_{i,j}\omu_{i,j}^2 = O(a_1)                                           &
	b_8 & = b_1.
\end{align*}
Then, the upper bound has the form
\begin{align*}
	\frac{1}{\sigma^3}\big[
	 & n^4(\underbrace{a_1^3}_{k^3/n^3} + \underbrace{a_1a_2}_{O(kS_2/n^4)} +\underbrace{a_3}_{O(S_3/n^4)} + \underbrace{a_4}_{O(L_4/n^4)})\underbrace{(b_1^3+b_1b_2+b_3+b_4)}_{O(1)} + \\
	 & n^3(\underbrace{a_5}_{O(S_2/n^3)} + \underbrace{a_1a_8}_{O(k^2/n^2)})\underbrace{(b_5+b_1b_8)}_{O(1)} + n^2\underbrace{a_6}_{O(k/n)}\underbrace{b_6}_{O(1)} \big],
\end{align*}
which can be simplified to
\[
	O(\frac{1}{\sigma^3}\big[
		nk^3 + kS_2 + S_3 + L_4 + S_2 + nk^2 + k/n
		\big]) = O\big(\frac{1}{\sigma^3}(nk^3 + kS_2 + S_3 + L_4)\big),
\]
which is what is claimed in the theorem.
\section{Experiments}\label{app:experiments}
\subsection{MMD}
\begin{figure*}[h]
	\begin{subfigure}[t]{0.32\linewidth}
		\includegraphics[width=\linewidth]{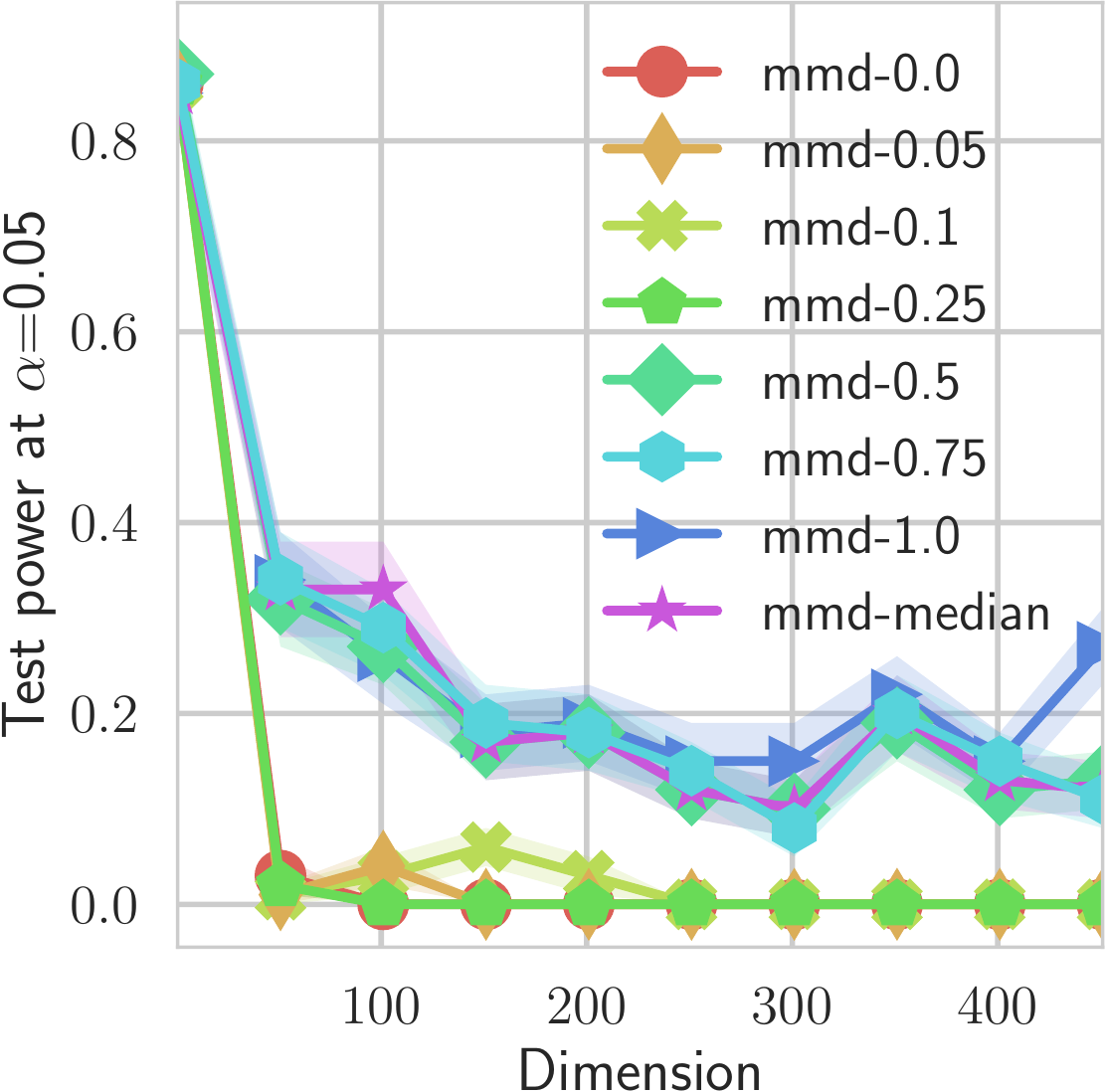}
		\caption{$\mu=0.5,\sigma=1,n_1=128$.}
	\end{subfigure}
	\begin{subfigure}[t]{0.32\linewidth}
		\includegraphics[width=\linewidth]{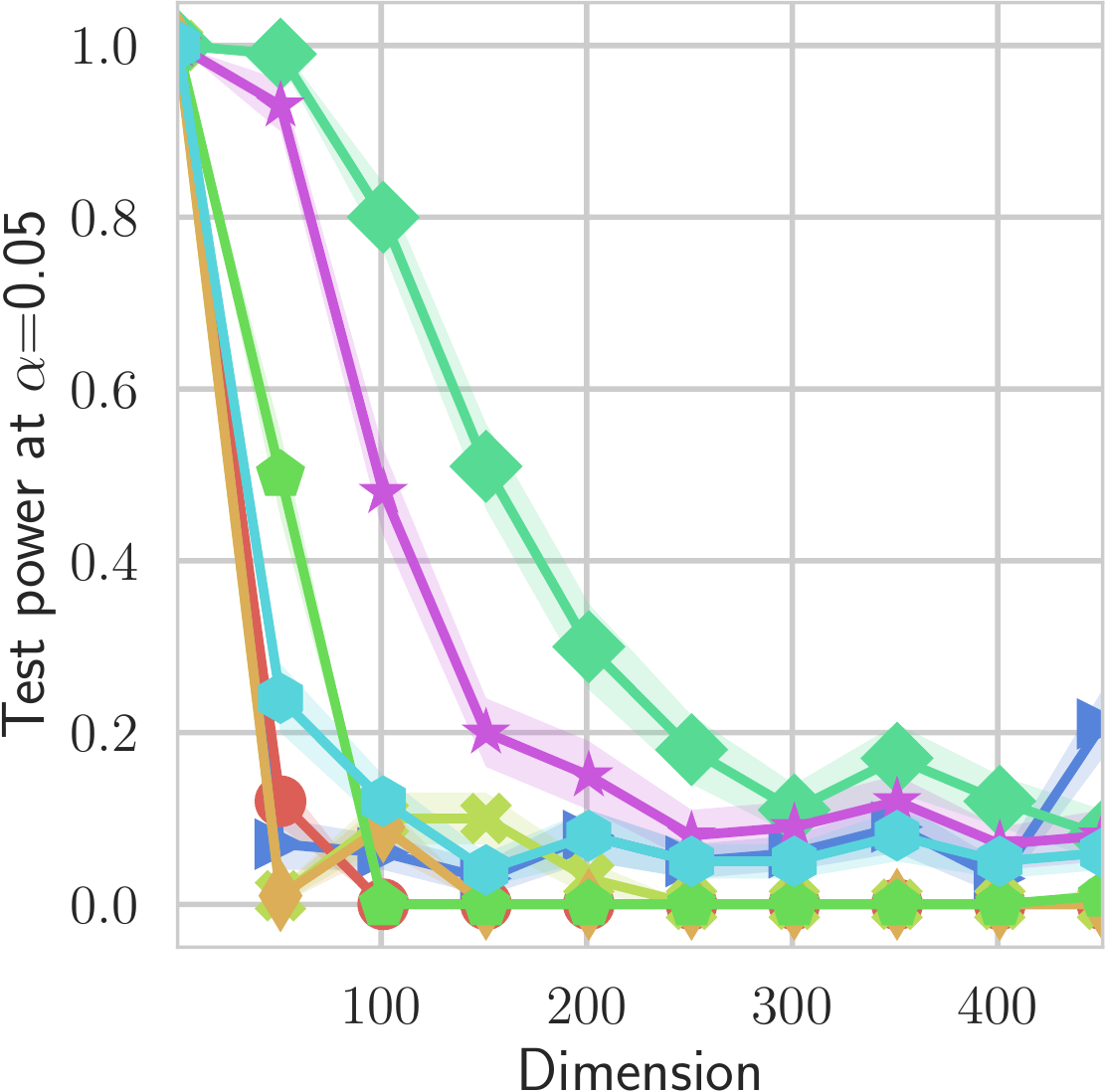}
		\caption{$\mu=0,\sigma=3,n_1=128$.}
	\end{subfigure}
	\begin{subfigure}[t]{0.32\linewidth}
		\includegraphics[width=\linewidth]{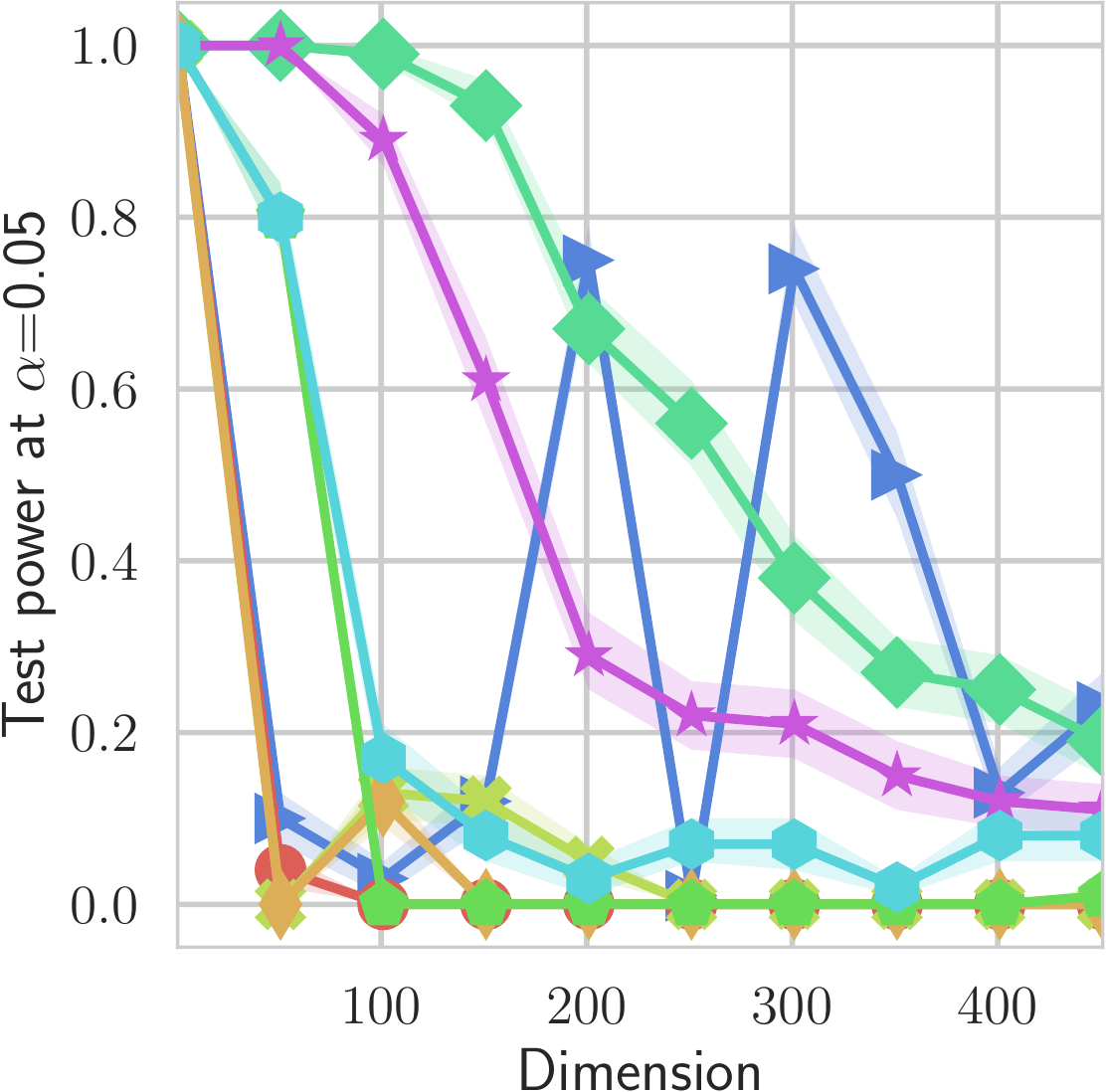}
		\caption{$\mu=0,\sigma=3,n_1=256$.}
	\end{subfigure}
	\caption{The different MMD tests on the three setups in \Cref{fig:power}. The legend is consistent across the panels.}
\end{figure*}

\newpage

\subsection{Architecture}
We have used the same architecture as in~\cite{li2015generative,sutherland2016generative}, which using the modules from \texttt{PyTorch} can be written as follows.
\begin{verbatim}
nn.Sequential(
    nn.Linear(noise_dim, 64),
    nn.ReLU(),
    nn.Linear(64, 256),
    nn.ReLU(),
    nn.Linear(256, 256),
    nn.ReLU(),
    nn.Linear(256, 1024),
    nn.ReLU(),
    nn.Linear(1024, ambient_dim))
\end{verbatim}
For MNIST we have also added a terminal \texttt{nn.Tanh} layer.

\subsection{Data}
We have used the MNIST data as packaged by \texttt{torchvision}, with the additional processing of scaling the output to $[-1, 1]$ as we are using a final \texttt{Tanh} layer.
For the \emph{two moons} data, we have used a noise level of $0.05$.

\subsection{Optimization}
All details are provided in the table below.
In some cases we have optimized with a larger step for a number of epochs, and then reduced it for the remaining epochs --- in the table below these are separated by commas.

\begin{center}
	\begin{tabular}{l | lll}
		\toprule
		Model              & Step size          & Batch size & Epochs   \\
		\midrule
		\Cref{fig:moons-a} & $10^{-4}$          & 256        & 500      \\
		\Cref{fig:moons-b} & $10^{-4}$          & 256        & 500      \\
		\Cref{fig:moons-c} & $10^{-4}$          & 256        & 500      \\
		\Cref{fig:mnist-a} & $10^{-3}, 10^{-4}$ & 256        & 500, 500 \\
		\Cref{fig:mnist-b} & $10^{-3}, 10^{-4}$ & 512        & 500, 500 \\
		\Cref{fig:mnist-c} & $10^{-3}, 10^{-4}$ & 128        & 100, 100 \\
		\Cref{fig:mnist-d} & $10^{-4}, 10^{-4}$ & 128        & 100, 100 \\
		\bottomrule
	\end{tabular}
\end{center}

\end{document}